\documentclass[10pt,twocolumn,letterpaper]{article}

\usepackage[pagenumbers]{cvpr} 

\usepackage[dvipsnames,table]{xcolor}

\usepackage{float}
\restylefloat{table}
\usepackage{adjustbox}
\usepackage{tabularray}
\UseTblrLibrary{booktabs}
\usepackage{booktabs}
\usepackage{diagbox}
\usepackage{bm}
\usepackage{caption}
\usepackage[percent]{overpic}

\SetTblrInner{rowsep=1pt}

\definecolor{cvprblue}{rgb}{0.21,0.49,0.74}
\usepackage[pagebackref,breaklinks,colorlinks,citecolor=cvprblue]{hyperref}

\def\paperID{895} 
\def\confName{CVPR}
\def\confYear{2024}

\title{Global Latent Neural Rendering}

\author{Thomas Tanay \quad\quad Matteo Maggioni\\
Huawei Noah's Ark Lab\vspace{0.1cm}\\
{\small Project page: \href{https://thomas-tanay.github.io/convglr}{\tt thomas-tanay.github.io/convglr}}
}

\makeatletter
\renewcommand{\paragraph}{%
  \@startsection{paragraph}{4}%
  {\z@}{0.8ex \@plus 0.2ex \@minus .2ex}{-1em}%
  {\normalfont\normalsize\bfseries}%
}
\makeatother

\begin{document}

\twocolumn[{%
\renewcommand\twocolumn[1][]{#1}%
\maketitle
\vspace{-0.4cm}
\begin{center}
    \centering
    \captionsetup{type=figure}
    \makebox[0.01\textwidth]{}\hfill
    \makebox[0.195\textwidth]{\scriptsize RegNeRF~\cite{niemeyer2022regnerf}}\hfill
    \makebox[0.195\textwidth]{\scriptsize SparseNeRF~\cite{wang2023sparsenerf}}\hfill
    \makebox[0.195\textwidth]{\scriptsize GPNR~\cite{suhail2022generalizable}}\hfill
    \makebox[0.195\textwidth]{\scriptsize GeoNeRF~\cite{johari2022geonerf}}\hfill
    \makebox[0.195\textwidth]{\scriptsize Challenge winner~\cite{jang2023vschh}}

    \makebox[0.01\textwidth]{\hspace{-0.2cm} \rotatebox{90}{\hspace{0.7cm} \textbf{baselines}}}\hfill
    \includegraphics[width=0.195\textwidth]{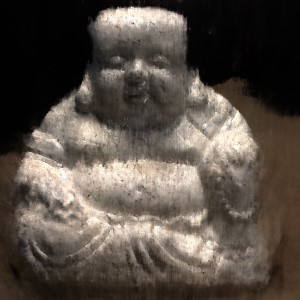}\hfill
    \includegraphics[width=0.195\textwidth]{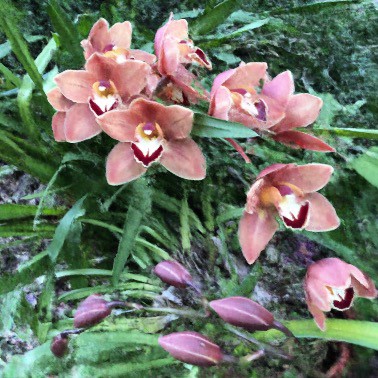}\hfill
    \includegraphics[width=0.195\textwidth]{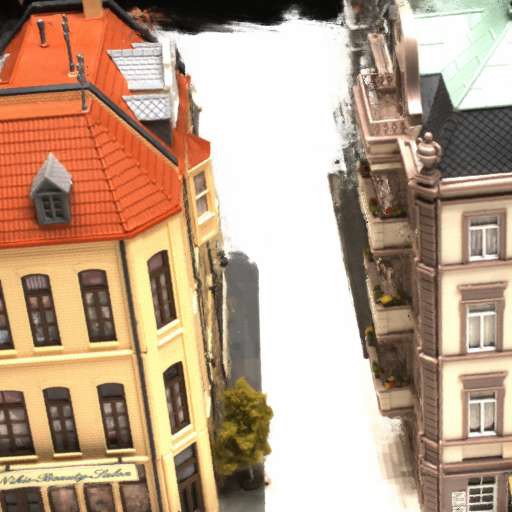}\hfill
    \includegraphics[width=0.195\textwidth]{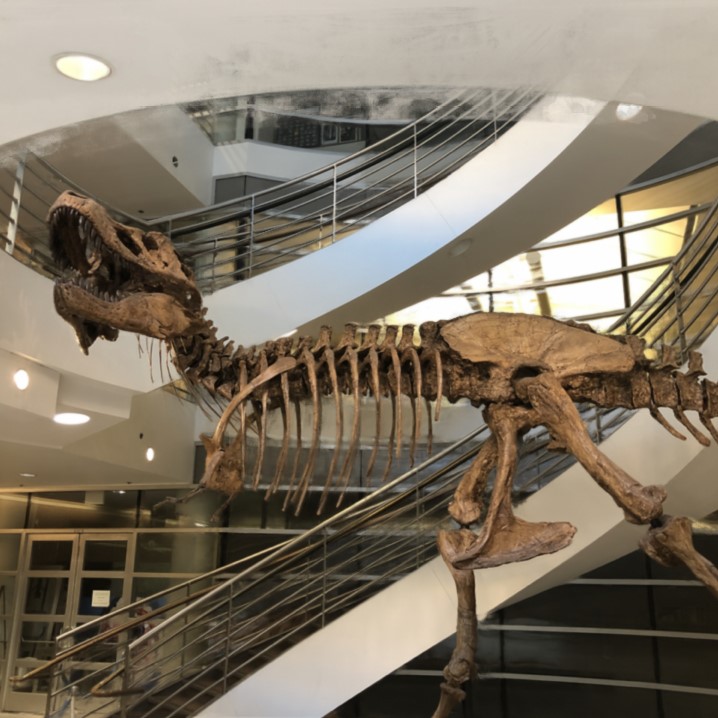}\hfill
    \includegraphics[width=0.195\textwidth]{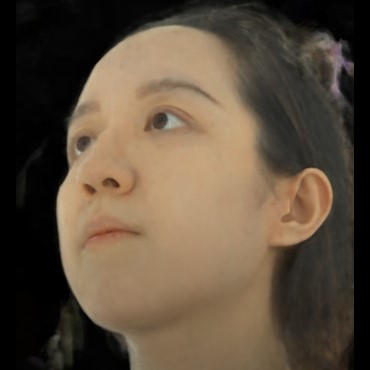}

    \makebox[0.01\textwidth]{\hspace{-0.2cm} \rotatebox{90}{\hspace{1.2cm} \textbf{ours}}}\hfill
    \includegraphics[width=0.195\textwidth]{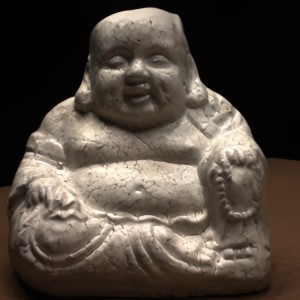}\hfill
    \includegraphics[width=0.195\textwidth]{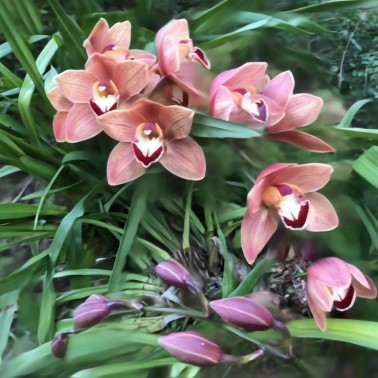}\hfill
    \includegraphics[width=0.195\textwidth]{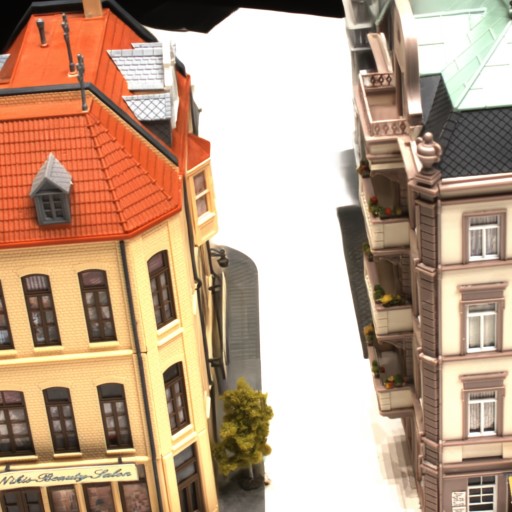}\hfill
    \includegraphics[width=0.195\textwidth]{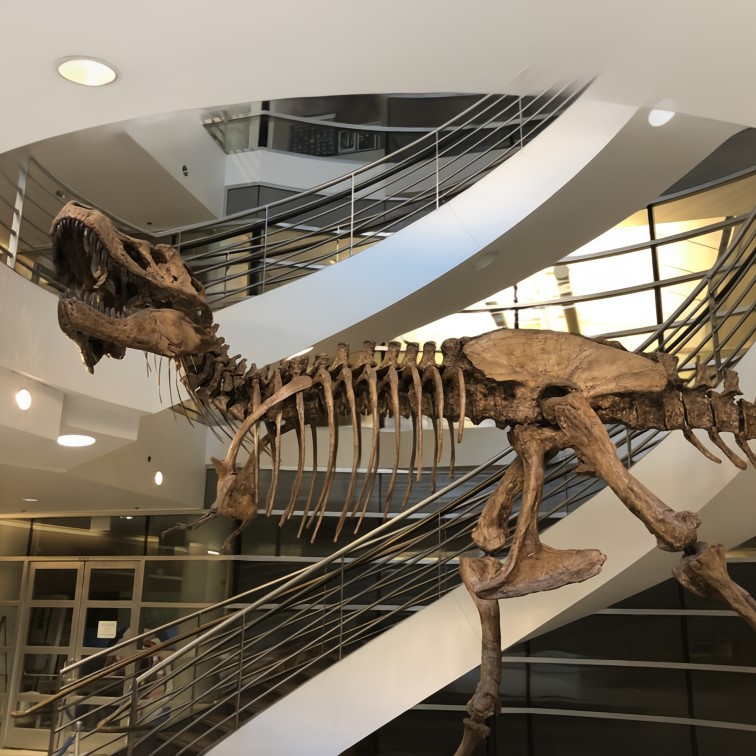}\hfill
    \includegraphics[width=0.195\textwidth]{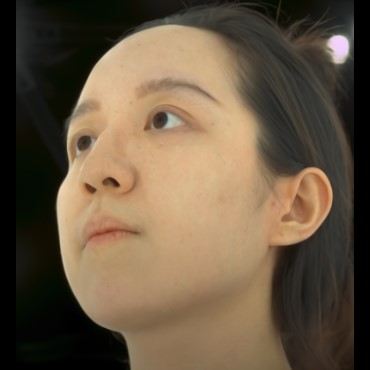}\vspace{-0.1cm}

    \makebox[0.01\textwidth]{}\hfill
    \makebox[0.195\textwidth]{\footnotesize \textbf{Sparse DTU}}\hfill
    \makebox[0.195\textwidth]{\footnotesize \textbf{Sparse RFF}}\hfill
    \makebox[0.195\textwidth]{\footnotesize \textbf{Generalizable DTU}}\hfill
    \makebox[0.195\textwidth]{\footnotesize \textbf{Generalizable RFF}}\hfill
    \makebox[0.195\textwidth]{\footnotesize \textbf{ILSH}}\vspace{-0.1cm}

    \makebox[0.01\textwidth]{}\hfill
    \makebox[0.195\textwidth]{\footnotesize 3 views}\hfill
    \makebox[0.195\textwidth]{\footnotesize 3 views}\hfill
    \makebox[0.195\textwidth]{\footnotesize unknown scene}\hfill
    \makebox[0.195\textwidth]{\footnotesize unknown scene}\hfill
    \makebox[0.195\textwidth]{\footnotesize ICCV 23 challenge}

   \vspace{-0.1cm}
    \captionof{figure}{Qualitative comparison of our method with various baselines under 5 different experimental setups. Our method renders target views in a low-resolution latent space and operates over all camera rays jointly. It produces significantly better geometries and textures than previous sparse and generalizable methods, which render light rays independently and typically suffer from grainy artifacts.}
    \label{fig:qualitative_comparison}
\end{center}%
}]

\begin{abstract}
A recent trend among generalizable novel view synthesis methods is to learn a rendering operator acting over single camera rays.
This approach is promising because it removes the need for explicit volumetric rendering, but it effectively treats target images as collections of independent pixels.
Here, we propose to learn a global rendering operator acting over all camera rays jointly.
We show that the right representation to enable such rendering is a \mbox{5-dimensional} plane sweep volume consisting of the projection of the input images on a set of planes facing the target camera. 
Based on this understanding, we introduce our Convolutional Global Latent Renderer \mbox{(ConvGLR)}, an efficient convolutional architecture that performs the rendering operation globally in a low-resolution latent space.
Experiments on various datasets under sparse and generalizable setups show that our approach consistently outperforms existing methods by significant margins.
\end{abstract}

\section{Introduction}
\label{sec:intro}

Significant progress has been made on novel view synthesis in recent years, both in terms of image quality and rendering speed~\cite{mildenhall2020nerf,barron2021mip,barron2023zipnerf,fridovich2022plenoxels,muller2022instant,chen2022tensorf,kerbl20233d}. 
However, a lot of this progress has focused on the scene-specific formulation of the problem, where models are trained to fit one scene.
We are interested here in the generalizable formulation, where novel views of unknown scenes can be rendered directly from a set of posed input views~\cite{mildenhall2019local,yu2021pixelnerf,wang2021ibrnet,chen2021mvsnerf,suhail2022generalizable}.

This generalizable formulation is challenging because it requires to reason about the geometry of the scene for each target image, instead of solving the geometry problem as a preliminary step. 
It also typically relies on a much sparser number of input views (3 to 16 here) while the scene-specific formulation routinely uses 100s of input views.
However, we believe that it is ultimately more powerful because 
1)~sparse setups are common in real world applications~\cite{niemeyer2022regnerf, deng2022depth, wang2023sparsenerf} and 
2)~it provides the ability to reason about unkown environments and could pave the way for the training of large scale 3D vision models~\cite{du2023cross}.
Most recent works on generalizable novel view synthesis learn to predict 5D radiance fields based on some form of geometric reasoning before applying volumetric rendering; a fixed operation consisting in integrating the radiance over light rays~\cite{yu2021pixelnerf,wang2021ibrnet,chen2021mvsnerf,johari2022geonerf}.
A recent development is to use a 4D light field approach and predict the color of camera rays directly, effectively learning the rendering operation itself~\cite{suhail2022light,suhail2022generalizable,du2023cross}. 
This later approach is promising because it removes the need for explicit volumetric rendering but so far, it is still implemented on a single-ray basis. 

In this work, we learn a global rendering operator acting over all camera rays jointly.
We achieve this by revisiting plane sweep volumes (PSVs), obtained by projecting the input views on a set of planes distributed parallel to the target image plane.
In particular, we observe that PSVs implicitly encode the epipolar geometry of the scene such that 
mixing information \emph{across epipolar lines} can be implemented with operations along the view dimension of PSVs, 
mixing information \emph{along epipolar lines} can be implemented with operations along the depth dimension of PSVs and 
mixing information \emph{between light rays} can be implemented with operations along the height and width dimensions of PSVs.
Based on this understanding, we introduce a Convolutional Global Latent Renderer \mbox{(ConvGLR)}, an efficient convolutional architecture that renders novel views directly from plane sweep volumes.
ConvGLR  is a 4 step model that 1)~arranges the PSV into groups of successive depths, 2)~aggregates information across views in a depth-independent manner while reducing the spatial dimension of the representation, 3)~performs global latent rendering by progressively collapsing the depth dimension and 4)~upsamples the rendered representation into a final output. This design is validated in mutliple experiments on the DTU~\cite{jensen2014large}, Real-Forward Facing~\cite{mildenhall2020nerf} and Spaces~\cite{flynn2019deepview} datasets under established sparse and generalizable setups~\cite{niemeyer2022regnerf,chen2021mvsnerf,mildenhall2020nerf}, as well as on the recently introduced ILSH dataset~\cite{zheng2023ilsh} in the context of a public novel view synthesis challenge with held-out test views~\cite{jang2023vschh,tonerfornottonerf}. 
Our main contributions are as follow:
\begin{itemize}[leftmargin=*,noitemsep,topsep=0pt]
\item We introduce \emph{global latent neural rendering}, a simple and generalizable approach to novel view synthesis that directly renders novel views from plane sweep volumes.
\item We design a \emph{Convolutional Global Latent Renderer} \mbox{(ConvGLR)}, a convolutional architecture that implements global latent neural rendering efficiently.
\item We evaluate ConvGLR extensively on sparse and generalizable setups as well as on a public novel view synthesis challenge with held-out test views, and significantly outperform existing methods in all cases.
\end{itemize}

\section{Related work}
\label{sec:related work}

\paragraph{NeRFs}
Neural Radiance Fields~\cite{mildenhall2020nerf,barron2021mip,barron2023zipnerf} model the 5D radiance and 3D density fields of individual scenes in the weights of an MLP. 
They have become highly popular for their ability to produce high quality renderings of complex scenes from arbitrary viewpoints.
They tend to be relatively slow at rendering time, although significant speed-ups have been obtained by removing the neural representation entirely~\cite{fridovich2022plenoxels}, using multiresolution hash encodings~\cite{muller2022instant}, tensor decompositions~\cite{chen2022tensorf} or 3D gaussians~\cite{kerbl20233d}.
NeRF models also struggle on scenes that are viewed under very sparse conditions. Multiple attempts have been made at addressing this limitation, often by training on missing views using auxiliary losses. 
For instance, DietNeRF~\cite{jain2021putting} uses a semantic consistency loss based on the CLIP vision transformer~\cite{radford2021learning}.
RegNeRF~\cite{niemeyer2022regnerf} uses appearance and geometry regularization based on a normalizing flow model and a smoothness loss.
FlipNeRF ~\cite{seo2023flipnerf} increases the number of training rays by reflecting the existing ones and introduces two new regularization losses.
MixNeRF ~\cite{seo2023mixnerf} models rays with mixture densities and introduces depth estimation as proxy objective.
DSNeRF~\cite{deng2022depth} exploits readily-available depth supervision signals obtained from COLMAP~\cite{schonberger2016structure}.
SparseNeRF~\cite{wang2023sparsenerf} improves the use of depth maps further by introducing a depth ranking constraint.
Similarly to our approach, GANeRF~\cite{roessle2023ganerf} improves the rendering operation by acting on groups of pixels via an adversarial loss applied on patches. 
However, this is in the context of a scene-specific model that still relies on fixed volumetric rendering over individual camera rays.

\paragraph{Light fields}
In free space, the radiance is constant over light rays and scenes can be encoded as 4D light fields.
This idea has been used in early works to perform novel view synthesis without~\cite{levoy1996light}, or with limited~\cite{gortler1996lumigraph} geometric reasoning by relying on a dense sampling of the scene.
Recent methods have focused on sparser setups in a learning based way~\cite{kalantari2016learning,sitzmann2021light,attal2022learning,suhail2022light,suhail2022generalizable},
often with a focus on modeling non-Lambertian effects~\cite{attal2022learning,suhail2022light}.
An important distinction between these works and neural radiance fields is that they learn the rendering operation instead of relying on classival volumetric rendering.
Contrary to our method, however, they still learn the rendering operation over single light rays.

\paragraph{Ray transformers}
A popular approach to novel view synthesis is to reason about the geometry of the scene implicitly~\cite{shum2000review}, typically via known epipolar constraints.
For instance, GRF~\cite{trevithick2021grf} and PixelNeRF~\cite{yu2021pixelnerf} extract image features along epipolar lines to encode 3D points, and render camera rays using volumetric rendering.
\mbox{IBRNet}~\cite{wang2021ibrnet} and NerFormer~\cite{reizenstein2021common} follow a similar approach while using more sophisticated transformer-based architectures. 
\mbox{DynIBaR}~\cite{li2023dynibar} extends epipolar line sampling in a motion-aware fashion.
LFNR~\cite{suhail2022light}, GPNR~\cite{suhail2022generalizable} and GNT~\cite{t2023is} also process image patches extracted along epipolar lines with transformers, but predict the color of individual camera rays directly without explicit volumetric rendering.
Finally, the method from~\cite{du2023cross} extends this approach to the challenging scenario of wide-baseline stereo pairs.
Our method also uses implicit geometric reasoning, but it does so with plane sweep volumes which are richer epipolar encodings than simple epipolar lines.

\paragraph{Explicit geometry}
In contrast with the previous category, a number of novel view synthesis methods rely on explicit geometric modeling of the scene~\cite{shum2000review}.
Early methods included 3D warping based on depth information~\cite{mcmillan1997image}, layered depth images to deal with occlusions~\cite{shade1998layered} or view-dependent texture maps inspired from computer graphics~\cite{debevec1996modeling}.
More recent methods still rely on depth maps~\cite{prinzler2023diner,deng2022depth,wang2023sparsenerf,Soft3DReconstruction} or rely on the construction of a geometric scaffold or mesh~\cite{hedman2017casual,riegler2020free,riegler2021stable,chibane2021stereo}.
However, these methods are vulnerable to inacuracies in the estimation of the underlying geometry. 
In contrast, our method does not use any form of explicit geometric reasoning.

\paragraph{Multiplane images}
The plane sweep algorithm was introduced in the context of multi-view stereo in~\cite{collins1996space} and was first applied to novel view synthesis using a layered representation in~\cite{szeliski1998stereo}.  
With the advent of deep learning, several methods have been introduced to perform generalizable novel view synthesis by processing PSVs.
Early methods typically produced layered representations that consisted in a mix of depth maps, oclusion maps and color maps~\cite{flynn2016deepstereo,Soft3DReconstruction,kalantari2016learning}.
Later methods focused on the multiplane image representation (MPI), which consists in a set of RGB$\alpha$ images that can be projected to novel viewpoints and rendered using alpha blending~\cite{zhou2018stereo,srinivasan2019pushing,flynn2019deepview,mildenhall2019local}.
MPIs have also been used in a scene-specific manner~\cite{wizadwongsa2021nex} and to generate novel views from a single image~\cite{tucker2020single,li2021mine,han2022single}.
Layered depth images are MPI variants where an extra depth channel is predicted~\cite{shade1998layered,lin2020deep,hu2021worldsheet,khakhulin2022stereo,solovev2023self}.
Finally, multiplane feature representations were recently introduced for multi-frame denoising~\cite{tanay2023efficient}.
Our method differs from these works in one important way: instead of producing a layered representation that is rendered through summation or alpha blending, it learns the rendering operation in a low-dimensional latent space.

\paragraph{3D cost volumes}
A variant of the plane sweep algorithm consists in extracting deep features from the input images indepedently, constructing plane sweep volumes from the deep features, and computing the variance over the input views~\cite{yao2018mvsnet}.
Such 3D cost volumes have been used extensively in the literature on multi-view stereo (MVS)~\cite{yao2019recurrent,im2019dpsnet,cheng2020deep,gu2020cascade,xu2020learning,yang2020cost}, and have recently been combined with NeRFs for novel view synthesis~\cite{chen2021mvsnerf,johari2022geonerf,lin2022efficient,liu2022neuray}.
MVSNeRF~\cite{chen2021mvsnerf} in particular computes a cost volume centered on the reference view, refines it with a 3D CNN, predicts radiance and density fields using an MLP and finally integrates over camera rays using volumetric rendering. 
GeoNeRF~\cite{johari2022geonerf} instead computes cascaded cost volumes centered on the input views, refines these cost volumes using multi-head attention, and again predicts radiance and density fields using MLPs before integrating over camera rays.
Our methods differs from these works in three ways: it uses a PSV representation instead of a cost volume, it learns the rendering operation instead of applying fixed volumetric rendering, and it renders all the camera rays jointly instead of independently.

\section{Background}
\label{sec:background}

Consider a set of $V$ \emph{input views} of a scene, consisting of color images and camera parameters. The images are of height $H$ and width $W$, with red-green-blue color channels, and can be stacked into a 4D tensor $\bm{I} \in \mathbb{R}^{V \times 3 \times H \times W}$. The camera parameters $\bm{P}$ consist of an intrinsic tensor $\bm{K} \in \mathbb{R}^{V \times 3 \times 3}$ and an extrinsic tensor that can be split into a rotation tensor~$\bm{R} \in \mathbb{R}^{V \times 3 \times 3}$ and a translation tensor~$\bm{t} \in \mathbb{R}^{V \times 3 \times 1}$. Now consider a distinct \emph{target view} with ground-truth image $\bm{I}_\ast\,$, and camera parameters  $\bm{P}_\ast = \{\bm{K}_\ast, \bm{R}_\ast, \,\bm{t}_\ast \}$. We are interested in \emph{novel view synthesis}, which consists in predicting an estimate $\bm{\tilde{I}}_\ast\,$ of the target image $\bm{I}_\ast\,$, given the input images $\bm{I}$, the input camera parameters $\bm{P}$ and the target camera parameters $\bm{P}_\ast$. 

There exists two main formulations of this problem. The first one learns a \emph{scene-specific} function $\mathcal{F}_{\bm{I}, \bm{P}}$ on the input views, such that novel views can be rendered from novel camera parameters:
$\bm{\tilde{I}}_\ast = \mathcal{F}_{\bm{I}, \bm{P}}(\bm{P}_\ast)$.
The function $\mathcal{F}_{\bm{I}, \bm{P}}$ is trained on views from a single pre-defined scene, and can be used to render novel views for that scene only.
The second formulation learns a \emph{scene-agnostic} or \emph{generalizable} function $\mathcal{F}$ on sets of input views and target camera parameters, such that novel views can be rendered from novel sets of input views and target camera parameters:
$\bm{\tilde{I}}_\ast = \mathcal{F}(\bm{I}, \bm{P}, \bm{P}_\ast)$. 
This time, the function $\mathcal{F}$ is trained on a large corpus of scenes, and can be used to render novel views from scenes that have not been seen during training.

The scene-specific formulation is a defining characteristic of NeRF~\cite{mildenhall2020nerf} and its extensions~\cite{barron2021mip,barron2023zipnerf,muller2022instant,chen2022tensorf} which model the function $\mathcal{F}_{\bm{I}, \bm{P}}$ indirectly through two fields: a radiance field returning a color for every point in space and viewing direction (5D$\to$ 3D function) and a density field returning a density for every point in space (3D$\to$1D function). The target image $\bm{\tilde{I}}_\ast$ is then rendered by integrating the two fields over camera rays using classical volumetric rendering. For scenes that mostly consist of free space (as is often the case), the 5D radiance field model is redundant because the radiance remains constant along light rays. Light field networks~\cite{sitzmann2021light,suhail2022light,attal2022learning} rely on this observation to directly model the function $\mathcal{F}_{\bm{I}, \bm{P}}$ as a light field returning a color for every light ray (4D$\to$3D function). 

Among generalizable methods, a well-known family are the models that predict multiplane image representations~\cite{zhou2018stereo, flynn2019deepview, srinivasan2019pushing, mildenhall2019local}. They typically process plane sweep volumes and predict a 3D radiance field with no view dependence (in their standard form) and a 3D density field as a discrete set of RGB$\alpha$ images, that are rendered through alpha-blending.
Generalizable neural radiance fields learn a NeRF model on top of a geometric representation, which can rely on 2D deep features extracted along epipolar lines~\cite{trevithick2021grf,yu2021pixelnerf,wang2021ibrnet} or 3D cost volumes~\cite{chen2021mvsnerf,johari2022geonerf,lin2022efficient,liu2022neuray}.
Existing generalizable light field networks~\cite{suhail2022generalizable,du2023cross} also extract image patches or features along epipolar lines, but they learn the rendering operation and directly predict a pixel color.
In this work, we introduce a generalizable light field model that learns to render images globally, by operating over all the camera rays jointly in a low-resolution latent space.
We summarize the difference between our approach and various previous methods in \Cref{table:method_comparison}.

\begin{table}[ht]
\centering
\begin{adjustbox}{width=\columnwidth}
\begin{tblr}{@{}l|ccc@{}}
\toprule[1.5pt]
methods & formulation & model & rendering \\
\hline\hline
\SetCell[r=2]{l} {Multi-plane images\\ \cite{zhou2018stereo, flynn2019deepview, srinivasan2019pushing, mildenhall2019local}} & \SetCell[r=2]{c} \textbf{generalizable} & \SetCell[r=2]{c} {3D radiance field\\ + 3D density field} & \SetCell[r=2]{c} {fixed\\ pointwise} \\ 
 & & & \\
\hline
\SetCell[r=2]{l} {Neural radiance fields\\ \cite{mildenhall2020nerf,barron2021mip,barron2023zipnerf,muller2022instant,chen2022tensorf}} & \SetCell[r=2]{c} scene-specific & \SetCell[r=2]{c} {5D radiance field\\ + 3D density field} & \SetCell[r=2]{c} {fixed\\ pointwise} \\ 
 & & & \\
\hline
\SetCell[r=2]{l} {Generalizable neural radiance\\ fields \cite{trevithick2021grf,yu2021pixelnerf,wang2021ibrnet,chen2021mvsnerf,johari2022geonerf,liu2022neuray}} & \SetCell[r=2]{c} \textbf{generalizable} & \SetCell[r=2]{c} {5D radiance field\\ + 3D density field} & \SetCell[r=2]{c} {fixed\\ pointwise} \\ 
 & & & \\
\hline
\SetCell[r=2]{l} {Light field networks\\ \cite{sitzmann2021light,suhail2022light,attal2022learning}} & \SetCell[r=2]{c} scene-specific & \SetCell[r=2]{c} \textbf{4D light field} & \SetCell[r=2]{c} {none / learned \\ pointwise} \\
 & & & \\
\hline
\SetCell[r=2]{l} {Generalizable light field\\ networks~\cite{suhail2022generalizable,t2023is,du2023cross}} & \SetCell[r=2]{c} \textbf{generalizable} & \SetCell[r=2]{c} \textbf{4D light field} & \SetCell[r=2]{c} {\textbf{learned}\\ pointwise} \\
 & & & \\
\hline
\SetCell[r=2]{l} {\textbf{Global latent}\\ \textbf{neural rendering (ours)}} & \SetCell[r=2]{c} \textbf{generalizable} & \SetCell[r=2]{c} \textbf{4D light field} & \SetCell[r=2]{c} {\textbf{learned}\\ \textbf{global}} \\
 & & & \\
\bottomrule[1.5pt]   
\end{tblr}
\end{adjustbox}
\caption{\textbf{Taxonomy of novel view synthesis approaches.} We distinguish methods according to the formulation they use (scene-specific vs generalizable), the model they learn (radiance field + density field vs light field) and the type of rendering they apply (fixed vs learned and pointwise vs global).}
\label{table:method_comparison}
\end{table}

\begin{figure*}[t]
  \centering
  \includegraphics[width=\linewidth]{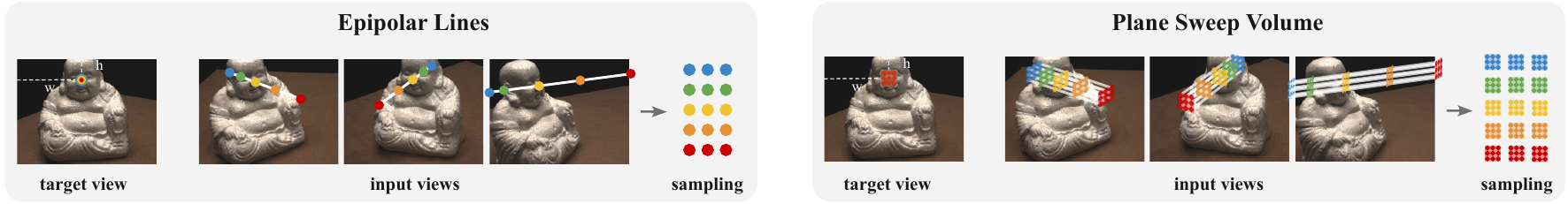}
   \caption{\textbf{Epipolar lines and the plane sweep volume.} 
Left: The camera ray passing through the pixel location $(h,w)$ in the target view projects as a set of epipolar lines in the input views.
Ray transformers~\cite{wang2021ibrnet,suhail2022generalizable,du2023cross,t2023is} process information sampled along these epipolar lines to predict the color of the target pixel $(h,w)$.
Right: The set of camera rays passing through adjacent pixel locations in the target view project as corresponding sets of epipolar lines in the input views.
Sampling along these sets of epipolar lines at constant depths defines a plane sweep volume facing the target view. Processing this plane sweep volume allows to render adjacent camera rays jointly.}
   \label{fig:epipolar_lines_psv}
\end{figure*}

\section{Method}
\label{sec:method}

We first define the Plane Sweep Volume (PSV) and highlight some of its interesting properties.
We then introduce global latent neural rendering, a new generalizable approach to novel view synthesis, and our Convolutional Global Latent Renderer (ConvGLR), an efficient implementation of it.
Finally we discuss some implementation details.

\subsection{The Plane Sweep Volume}

Consider a set of $D$ depth planes distributed parallel to the target image plane $\bm{I}_\ast\,$ such that they share the same normal~$\bm{n}_\ast$. The depth planes are uniquely defined by their distances $\{a_d\}_{d=1}^D$ from the target camera center and these distances are assumed to be chosen such that the scene of interest is adequately covered (we discuss the choice of these distances in practice in \cref{sec:implementation_details}). The plane sweep volume (PSV) is defined as the 5D tensor $\bm{X} \in \mathbb{R}^{D \times V \times 3 \times H \times W}$, obtained by projecting each input image $\bm{I}_v$ on each of the $D$ depth planes.\footnote{The term \emph{plane sweep volume} is often used to refer to 4D tensors obtained by projecting \emph{one} input view on the depth planes. The definition used here generalizes this to more views.} Formally, each projected image $\bm{X}_{dv}$ is obtained by applying a homography to $\bm{I}_v$, represented by a $3\!\times\!3$ matrix $\bm{H}_{dv}$. Assuming without loss of generality that the world origin is at the target camera center such that $\bm{R}_\ast$ is the identity, $\bm{t}_\ast = \bm{0}$ and $\bm{n}_\ast = (0, 0, 1)^\top$, each homography matrix is defined as~\cite{hartley2003multiple}:
$\bm{H}_{dv} = \bm{K}_v \left( \bm{R}_v - \frac{\bm{t}_v \, {\bm{n}_\ast}^\top}{a_d} \right) \bm{K}_\ast^{-1}$.

The plane sweep volume is a highly structured tensor that encodes the epipolar geometry between the input views and the target view~\cite{collins1996space,szeliski1998stereo,flynn2016deepstereo}. 
Indeed, consider the camera ray passing through a pixel location $(h,w)$ in the target image plane. 
This camera ray projects as a set of epipolar lines in the input views. Then by construction, the PSV slice:
$\bm{r}_{hw} = \{\{\{\bm{X}_{dvchw}\}_{c=1}^3\}_{d=1}^D\}_{v=1}^V$
contains pixels sampled along these epipolar lines at matching depths.
In other words, $\bm{r}_{hw}$ can be seen as an encoding of the camera ray passing through $(h,w)$, given the input views. 
This is useful, because adjacent camera rays have adjacent encodings in the PSV and can be processed together using simple local operators (see \cref{fig:epipolar_lines_psv} and Supplementary Material). 
More precisely, the PSV is structured such that 1) operations along the depth dimension are operations along individual epipolar lines, 2) operations along the view dimension are operations between corresponding epipolar lines and 3) operations along the height and width dimensions are operations between nearby camera rays.

\subsection{Global Latent Neural Rendering}

We propose a simple and powerful novel view synthesis approach that consists in learning a generalizable light field model $\mathcal{F}$, directly from plane sweep volumes: 
$\bm{\tilde{I}}_\ast = \mathcal{F}(\bm{X})$
where $\mathcal{F}$ is implemented as a convolutional neural network.
This approach fundamentally differs from the recent line of works that use transformers to process image patches extracted along epipolar lines~\cite{wang2021ibrnet,suhail2022light,suhail2022generalizable,liu2022neuray,li2023dynibar,du2023cross}, because it uses the plane sweep volume to organise the computation and allows to process camera rays jointly. It also differs from the line of works on layered representations and multiplane images~\cite{flynn2016deepstereo, Soft3DReconstruction, kalantari2016learning, zhou2018stereo, flynn2019deepview, mildenhall2019local}, because it learns the rendering operation, instead of keeping the depths separated and relying on alpha-compositing.

The main challenge faced by our proposed approach is the size of the PSV: a 5D tensor $\bm{X} \in \mathbb{R}^{D \times V \times 3 \times H \times W}$ needs to be processed efficiently using convolutions to produce a 3D rendered image $\bm{\tilde{I}}_\ast \in \mathbb{R}^{3 \times H \times W}$. Our solution is illustrated in \Cref{fig:ConvGLR_overview} and has the following structure (see the Supplementary Material for more details).

\begin{figure}[ht]
  \centering
  \includegraphics[width=\columnwidth]{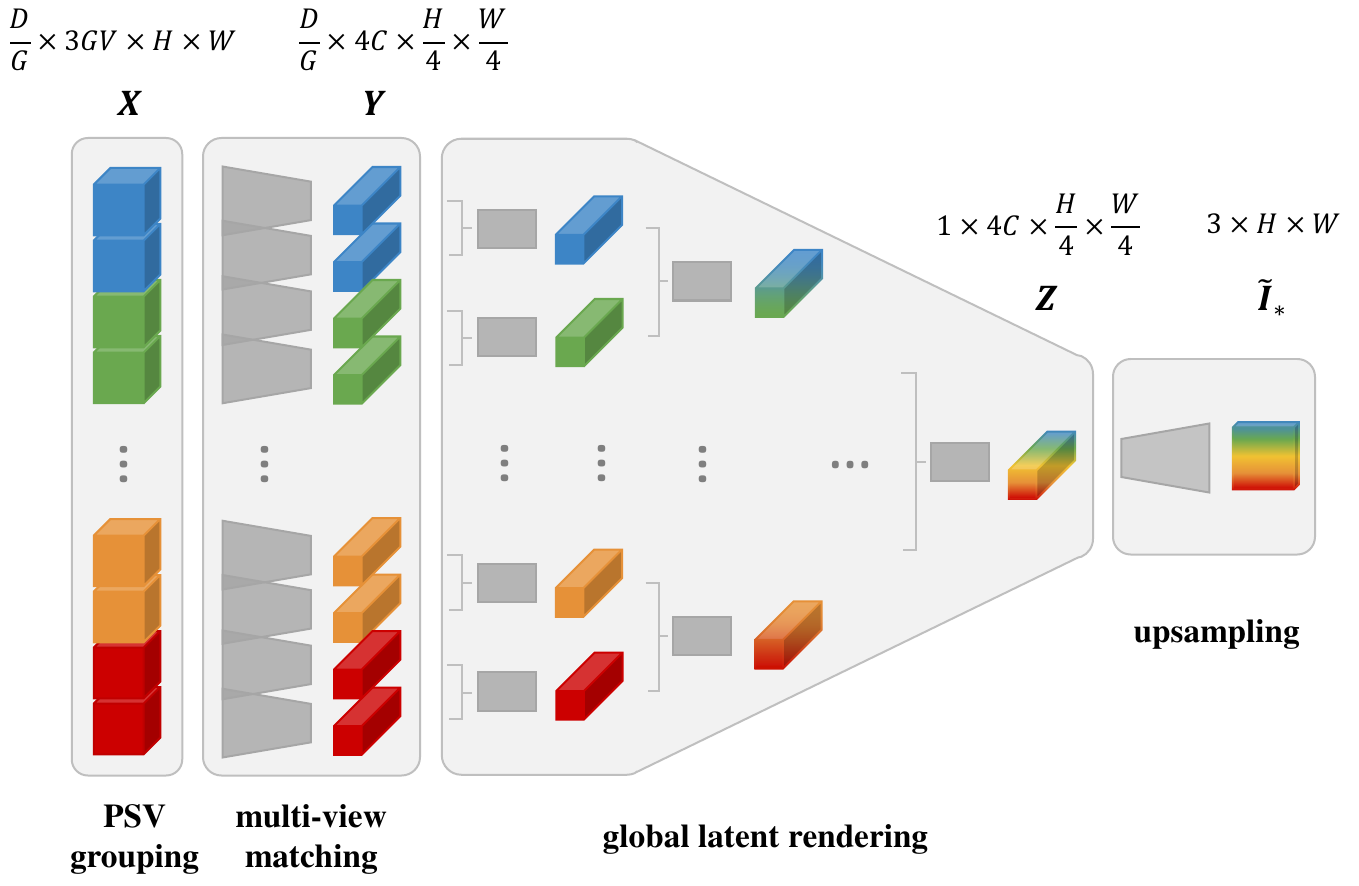}
   \caption{\textbf{Overview of ConvGLR.} The 4D grouped PSV $\bm{X}$ is turned into a latent volumetric representation $\bm{Y}$, then rendered into a latent novel view $\bm{Z}$ and finally upsampled into the novel view $\bm{\tilde{I}}_\ast$. All the dark gray blocks are implemented with 2D convolutions and resblocks. }
   \label{fig:ConvGLR_overview}
\end{figure}

\paragraph{PSV grouping} Similarly to the literature on multiplane representations~\cite{flynn2019deepview, mildenhall2019local}, the 5D PSV is treated as a 4D tensor of shape $D{\scriptstyle \times}3V{\scriptstyle \times}H{\scriptstyle \times}W$, such that the input views are processed together from the very first layer of the network. We show in our ablation study (see \cref{table:ablation}) that this approach is more powerful that the alternative one that constructs a 3D cost-volume~\cite{chen2021mvsnerf,johari2022geonerf,lin2022efficient,liu2022neuray}.
We then view the PSV as a tensor of shape $\frac{D}{G}{\scriptstyle \times}3GV{\scriptstyle \times}H{\scriptstyle \times}W$ for a group size $G$. This step significantly reduces the computational load by allowing to process the depths in groups, and effectively reduces the number of depths from $D$ to $D_{\scriptscriptstyle{\!G}} = \frac{D}{G}$.

\paragraph{Multi-view matching} Early layers aggregate information accross views, and treat the $D_{\scriptscriptstyle{\!G}}$ depths independently from each other by keeping them in the batch dimension. The spatial resolution is reduced 4$\times$, alternatively using 2D convolutions with stride 2 and 2D resblocks, following a typical encoder-decoder or Unet~\cite{ronneberger2015u} structure. The number of channels at the base of the network is a hyperparameter $C$, and the channels are doubled after each spatial downsampling. This block results in a latent volumetric representation $\bm{Y} \in \mathbb{R}^{D_{\scriptscriptstyle{\!G}} \times 4C \times \frac{H}{4} \times \frac{W}{4}}$.

\paragraph{Global latent rendering} The rendering operation is fundamentally an integration over the depth dimension, and consists in reducing the depth of the latent tensor $\bm{Y}$ to $1$. 
We implement it by iteratively grouping the depths by pairs and processing them with 2D resblocks. 
This emulates the use of 3D resblocks with a kernel size of 2 and a stride of 2 along the depth dimension, without requiring memory-expensive transpose operations.
This block produces a globally rendered latent representation $\bm{Z} \in \mathbb{R}^{1 \times 4C \times \frac{H}{4} \times \frac{W}{4}}$.

\paragraph{Upsampling} Finally, the output $\bm{\tilde{I}}_\ast$ is produced by upsampling the latent representation 4$\times$, alternatively using 2$\times$ bilinear interpolation and 2D resblocks, as is typically done in the super-resolution literature~\cite{lim2017enhanced,chan2021basicvsr}.

\begin{table*}[ht] 
\centering
\begin{adjustbox}{width=\textwidth}
\begin{tblr}{@{}l|c|cccccc|cccccc|cccccc@{}}
\toprule[1.5pt]
\SetCell[r=3]{c} Method & \SetCell[r=3]{c} Setting & \SetCell[c=6]{c}{PSNR$\uparrow$} &  &  &  &  &  & \SetCell[c=6]{c}{SSIM$\uparrow$} &  &  &  &  &  & \SetCell[c=6]{c}{LPIPS$\downarrow$} &  &  &  &  &  \\
\cmidrule[lr]{3-8} \cmidrule[lr]{9-14} \cmidrule[lr]{15-20}
 &  & \SetCell[c=2]{c}{3-view} & & \SetCell[c=2]{c}{6-view} & & \SetCell[c=2]{c}{9-view} & & \SetCell[c=2]{c}{3-view} & & \SetCell[c=2]{c}{6-view} & & \SetCell[c=2]{c}{9-view} & & \SetCell[c=2]{c}{3-view} & & \SetCell[c=2]{c}{6-view} & & \SetCell[c=2]{c}{9-view} \\
\cmidrule[lr]{3-4} \cmidrule[lr]{5-6} \cmidrule[lr]{7-8} \cmidrule[lr]{9-10} \cmidrule[lr]{11-12} \cmidrule[lr]{13-14} \cmidrule[lr]{15-16} \cmidrule[lr]{17-18} \cmidrule[lr]{19-20}
 &  & full & masked & full & masked & full & masked & full & masked & full & masked & full & masked & full & masked & full & masked & full & masked \\
\hline\hline
SRF~\cite{chibane2021stereo} & \SetCell[r=4]{c} {generalizable\\ unknown scene} & 15.84 & 15.32 & 17.77 & 17.54 & 18.56 & 18.35 & 0.532 & 0.671 & 0.616 & 0.730 & 0.652 & 0.752 & 0.482 & 0.304 & 0.401 & 0.250 & 0.359 & 0.232  \\
PixelNeRF~\cite{yu2021pixelnerf} &  & \SetCell{yellow!25} 18.74 & 16.82 & 21.02 & 19.11 & 22.23 & 20.40 & 0.618 & 0.695 & 0.684 & 0.745 & 0.714 & 0.768 & 0.401 & 0.270 & 0.340 & 0.232 & 0.323 & 0.220\\
MVSNeRF~\cite{chen2021mvsnerf} &  & 16.33 & 18.63 &  18.26 & 20.70 & 20.32 & 22.40 & 0.602 & \SetCell{yellow!25} 0.769 & 0.695 & 0.823 & 0.735 & 0.853 & 0.385 & 0.197 & 0.321 & 0.156 & 0.280 & 0.135  \\
\textbf{ConvGLR (Ours)} &  & \SetCell{orange!25} 20.47 & \SetCell{orange!25} 21.57 & \SetCell{orange!25} 25.23 & \SetCell{orange!25} 23.76 & \SetCell{orange!25} 26.98 & \SetCell{orange!25} 25.44 & \SetCell{orange!25} 0.784 & \SetCell{orange!25} 0.846 & \SetCell{orange!25} 0.843 & \SetCell{orange!25} 0.878 & \SetCell{orange!25} 0.878 & \SetCell{orange!25} 0.907 & \SetCell{orange!25} 0.249 & \SetCell{orange!25} 0.159 & \SetCell{orange!25} 0.189 & \SetCell{yellow!25} 0.123 & \SetCell{orange!25} 0.147 & \SetCell{yellow!25} 0.090 \\
\hline
SRF ft~\cite{chibane2021stereo} & \SetCell[r=4]{c} {generalizable\\ known scene} & 16.06 & 15.68  & 18.69 & 18.87  & 19.97 & 20.75  & 0.550 & 0.698 & 0.657 & 0.757 & 0.678 & 0.785 & 0.431 & 0.281 & 0.353 & 0.225 & 0.325 & 0.205\\
PixelNeRF ft~\cite{yu2021pixelnerf} &  & 17.38 & 18.95 & \SetCell{yellow!25} 21.52 & 20.56  & 21.67 & 21.83  & 0.548 & 0.710  & 0.670 & 0.753 & 0.680 & 0.781 & 0.456 & 0.269 & 0.351 & 0.223 & 0.338 & 0.203 \\
MVSNeRF ft~\cite{chen2021mvsnerf} &  & 16.26 & 18.54 & 18.22 & 20.49 & 20.32 & 22.22 & 0.601 & \SetCell{yellow!25} 0.769 & 0.694 & 0.822 & 0.736 & 0.853 & 0.384 & 0.197 & 0.319 & 0.155 & 0.278 & 0.135\\
\textbf{ConvGLR ft (Ours)} &  & \SetCell{red!25} 20.52 & \SetCell{red!25} 21.80 & \SetCell{red!25} 25.48 & \SetCell{red!25} 24.13 & \SetCell{red!25} 27.31 & \SetCell{red!25} 25.85 & \SetCell{red!25} 0.790 & \SetCell{red!25} 0.853 & \SetCell{red!25} 0.852 & \SetCell{red!25} 0.886 & \SetCell{red!25} 0.883 & \SetCell{red!25} 0.911 & \SetCell{red!25} 0.237 & \SetCell{red!25} 0.147 & \SetCell{red!25} 0.175 & \SetCell{red!25} 0.110 & \SetCell{red!25} 0.139 & \SetCell{red!25} 0.084 \\
\hline
mip-NeRF~\cite{barron2021mip} & \SetCell[r=7]{c} {scene-specific} & 7.64 & 8.68 & 14.33 & 16.54  & 20.71 & 23.58 & 0.227 & 0.571  & 0.568 & 0.741 & 0.799 & 0.879 & 0.655 & 0.353 & 0.394 & 0.198 & 0.209 & 0.092\\
DietNeRF~\cite{jain2021putting} &  & 10.01 & 11.85 & 18.70 & 20.63  & 22.16 & 23.83 & 0.354 & 0.633 &  0.668 & 0.778 & 0.740 & 0.823 & 0.574 & 0.314 & 0.336 & 0.201 & 0.277 & 0.173\\
RegNeRF~\cite{niemeyer2022regnerf} &  & 15.33 & 18.89 & 19.10 & 22.20 & \SetCell{yellow!25} 22.30 & 24.93 & \SetCell{yellow!25} 0.621 & 0.745 & \SetCell{yellow!25} 0.757 & \SetCell{yellow!25} 0.841 & \SetCell{yellow!25} 0.823 & \SetCell{yellow!25} 0.884 & \SetCell{yellow!25} 0.341 & \SetCell{yellow!25} 0.190 & \SetCell{yellow!25} 0.233 & \SetCell{orange!25} 0.117 & \SetCell{yellow!25} 0.184 & \SetCell{orange!25} 0.089 \\
MixNeRF~\cite{seo2023mixnerf} &  & \SetCell{gray!10} &  18.95 & \SetCell{gray!10} & 22.30 & \SetCell{gray!10} & 25.03 & \SetCell{gray!10} & 0.744 & \SetCell{gray!10} & 0.835 & \SetCell{gray!10} & 0.879 & \SetCell[c=6]{gray!10} & & & & & \\
FlipNeRF~\cite{seo2023flipnerf} &  & \SetCell{gray!10} &  \SetCell{yellow!25} 19.55 & \SetCell{gray!10} & \SetCell{yellow!25} 22.45 & \SetCell{gray!10} & \SetCell{yellow!25} 25.12 & \SetCell{gray!10} & 0.767 & \SetCell{gray!10} & 0.839 & \SetCell{gray!10} & 0.882 & \SetCell[c=6]{gray!10} & & & & & \\
DSNeRF~\cite{deng2022depth} & \SetCell[r=2]{c} Depth guided  & 16.90 & \SetCell{gray!10} & 20.60 & \SetCell{gray!10} & \SetCell{yellow!25} 22.30 & \SetCell{gray!10} & 0.570 & \SetCell{gray!10} & 0.750 & \SetCell{gray!10} & 0.810 & \SetCell{gray!10} & \SetCell[c=6]{gray!10} & & & & & \\
SparseNeRF~\cite{wang2023sparsenerf} &  & \SetCell{gray!10} & \SetCell{yellow!25} 19.55 & \SetCell[c=4]{gray!10} &  &  &  & \SetCell{gray!10} & \SetCell{yellow!25} 0.769 & \SetCell[c=4]{gray!10} &  &  &  & \SetCell{gray!10} & 0.201 & \SetCell[c=4]{gray!10} &  & & \\
\bottomrule[1.5pt]
\end{tblr}
\end{adjustbox}
\caption{\textbf{Sparse DTU.} Scenarios with 3, 6 and 9 input views.  We reproduce the values reported by~\cite{niemeyer2022regnerf} for~\cite{chibane2021stereo,yu2021pixelnerf,chen2021mvsnerf,barron2021mip,jain2021putting,niemeyer2022regnerf} and the values reported by each for~\cite{seo2023mixnerf,seo2023flipnerf,deng2022depth,wang2023sparsenerf}. We do not reproduce the LPIPS values of~\cite{seo2023mixnerf,seo2023flipnerf,deng2022depth} as they were computed using the AlexNet variant of LPIPS. We also note that the values reported by~\cite{deng2022depth} were computed on the full images. When a value is not available in the original publication, we simply gray the cell out. For each metric, 1st, 2nd and 3rd best-performing methods are highlighted in red, orange and yellow respectively.}
\label{table:sparse_DTU}
\end{table*}

\subsection{Additional conditioning}

While the PSV is an information-rich encoding of the input views, we propose to augment it further with two additional conditional inputs.
We show in our ablation study (see \cref{table:ablation}) that these two conditional inputs have a negligible negative impact on the computational load, but have a significant positive impact on performance. 

\paragraph{Positional encoding} 
First, we concatenate to the PSV the spatial coordinates $(h,w)$ in the form of two extra channels normalized in the $[0,1]$ range. We do not use any Fourrier encoding to avoid overloading an already large PSV tensor. Explicitly feeding the spatial coordinates is a simple way to make the model \emph{spatially-adaptive}~\cite{liu2018intriguing}, such that it renders specific groups of pixels differently depending on their location in the image (e.g. outer pixels are more likely to be of specific colors).
This use of positional encoding is closer to its original use in transformers~\cite{vaswani2017attention}, where it was introduced as a way of injecting information about the position of tokens in a sequence, than its use in NeRF~\cite{mildenhall2020nerf}, where it helps encode high-frequency content. 

\paragraph{Angular encoding}
Let $\bm{u}_{dv}$ be the unit vector pointing in the direction between the camera center of view $v$ and the center of the depth plane $d$. Remembering that the normal to the target image plane is $\bm{n}_\ast$, we concatenate the dot product $\bm{u}_{dv} \cdot \bm{n}_\ast$ as an additional channel to each projected image $\bm{X}_{dv}$ in the PSV. The motivation is two-fold. First, this dot product measures an angular distance between the target view and view $v$ (as seen from the depth plane $d$), and hence, it is a good measure of the similarity between the two views at that depth.
Second, we hypothesise that this can help model finegrained view-dependent effects, by making the input more explicitly view-dependent.\footnote{However, we observe that the PSV is already view-dependent and the angular distance could be computed implicitly by measuring the magnitude of the translations between successive depths.}

\subsection{Implementation details}
\label{sec:implementation_details}

Similarly to other novel view synthesis methods, the \emph{near} and \emph{far} bounds are important hyperparameters that can have a big impact on the performance of the method.
For the experiments on the DTU dataset, we empirically chose a near bound of 0.85 and a far bound of 1.75 for all scenes and target viewpoints. 
For the experiments on the RFF, LLFF and IBRNet datasets, we follow the established practice of using the bounds determined by COLMAP~\cite{schonberger2016structure}, with 0.9 and 1.1 factors for the near and far bounds respectively.
More generally, the choice of distances $\{a_d\}_{d=1}^D$---which determines the distribution of depth planes in the scene---faces similar issues to the choise of sample points along rays in volumetric rendering.
While sophisticated sampling strategies exist~\cite{barron2023zipnerf}, we chose two standard distributions. 
We sample the distances uniformly in depth for DTU and ILSH, and uniformly in disparity for RFF, LLFF and IBRNet.
For the hyperparameters of the ConvGLR model, we used $D = 128$ and $G = 4$, corresponding to an effective number of depths $D_{\scriptscriptstyle{\!G}} = 32$, and $C = 128$ in all our experiments.
The model is relatively large with 40M parameters (95M when the parameters of the rendering blocks are not shared), but it is fast, rendering a 375×512 image in 0.71 seconds on a single GPU.
Unless stated otherwise, all our models are trained with the Adam optimizer for 120k steps with a learning rate of 1.5e-4, decreased to 1.5e-5 in the last 20\% of the training and optionally to 1.5e-6 for the last 5\%.
We train on patches of 360$\times$360 pixels (or full images for Sparse DTU) with a batch size of 4 or 8 depending on the experiment, using 4 or 8 GPUs respectively.
We use a standard VGG loss~\cite{zhou2018stereo, flynn2019deepview, mildenhall2019local}, which we switch to an L1 loss in the last 10\% of the training to avoid gridding artifacts.
We use gradient clipping to stabilize the training.

\section{Experiments}
\label{sec:experiments}

We evaluate our method under sparse and generalizable novel view synthesis scenarios.
We consider 5 different experimental setups, using 3 different validation datasets, as detailed below.
In all cases, our convolutional global latent renderer (ConvGLR) significantly outperforms the baselines.
Qualitative comparisons are available in \cref{fig:qualitative_comparison} and in the Supplementary Material. 

\paragraph{\Cref{table:sparse_DTU}: Sparse DTU} We reproduce the setup introduced in PixelNeRF~\cite{yu2021pixelnerf}, refined in RegNeRF~\cite{niemeyer2022regnerf} and used in~\cite{seo2023mixnerf,seo2023flipnerf,deng2022depth,wang2023sparsenerf} on the DTU dataset~\cite{jensen2014large}.
In this setup, the images are downsampled 4$\times$ to a resolution of 400$\times$300. Images with incorrect exposure are excluded.\footnote{Images [3, 4, 5, 6, 7, 16, 17, 18, 19, 20, 21, 36, 37, 38, 39].}
The dataset is split into 88 scenes for training with 7 lighting conditions and 15 scenes for validation.\footnote{Scans [8, 21, 30, 31, 34, 38, 40, 41, 45, 55, 63, 82, 103, 110, 114].}
Three scenarios are considered with 3, 6 and 9 input views.\footnote{First 3, 6 and 9 images in [25, 22, 28, 40, 44, 48, 0, 8, 13].}
Validation is performed on all the views that are not input views or excluded views for all the validation scenes, with lighting condition nb.~3.
We report PSNR, SSIM and LPIPS (VGG variant) metrics computed on full and masked images, using object masks produced by~\cite{niemeyer2022regnerf}.
We train 3 different models with 3, 6 and 9 input views on the 88 training scenes (ConvGLR).
We see that they outperform all the baseline in all 3 scenarios by significant margins, especially on the full images due to a strong ability to generalize the background across scenes. 
We then finetune each model once on the input views of the 15 validation scenes for 10k steps (ConvGLR ft). To prevent the model from learning an identity function, we continue exposing it to training scenes, where the target views are distinct from the input views.
These models further improve their performances on novel views of the validation scenes.

\paragraph{\Cref{table:sparse_RFF}: Sparse RFF} 
We reproduce the setup introduced in RegNeRF~\cite{niemeyer2022regnerf} and used in~\cite{seo2023mixnerf,seo2023flipnerf,deng2022depth,wang2023sparsenerf} on the Real-Forward Facing dataset (RFF)~\cite{mildenhall2020nerf} for 3 input views.
In this setup, the images are downsampled 8$\times$ to a resolution of 504$\times$378.
Every 8th image is used for validation, and the 3 input views are selected evenly from the remaining images.
We report PSNR, SSIM and LPIPS (VGG) computed on full images.
While it was suggested in~\cite{niemeyer2022regnerf} that the LLFF dataset~\cite{mildenhall2019local} is too small for training generalizable methods (36 scenes), we found that finetuning a DTU trained model on LLFF provides good performance (ConvGLR).
Again, finetuning our model on the set of 8 validation scenes improves performance futher (ConvGLR ft).

\begin{table}[ht]
\centering
\begin{adjustbox}{width=0.7\columnwidth}
\begin{tblr}{@{}l|c|ccc@{}}
\toprule[1.5pt]
Method & Setting & PSNR$\uparrow$ & SSIM$\uparrow$ & LPIPS$\downarrow$ \\
\hline\hline
SRF~\cite{chibane2021stereo} & \SetCell[r=4]{c} {generalizable\\ unknown scene} & 12.34 & 0.250 & 0.591 \\
PixelNeRF~\cite{yu2021pixelnerf} & & 7.93 & 0.272 & 0.682 \\
MVSNeRF~\cite{chen2021mvsnerf} & & 17.25 & 0.557 & 0.356 \\
\textbf{ConvGLR (Ours)} &  & \SetCell{orange!25} 19.95 & \SetCell{orange!25} 0.700 & \SetCell{orange!25} 0.262 \\
\hline
SRF ft~\cite{chibane2021stereo} & \SetCell[r=4]{c} {generalizable\\ known scene} & 17.07 & 0.436 & 0.529  \\
PixelNeRF ft~\cite{yu2021pixelnerf} &  & 16.17 & 0.438 & 0.512 \\
MVSNeRF ft~\cite{chen2021mvsnerf} &  & 17.88 & 0.584 & \SetCell{yellow!25} 0.327 \\
\textbf{ConvGLR ft (Ours)} &  & \SetCell{red!25} 20.53 & \SetCell{red!25} 0.711 & \SetCell{red!25}  0.253 \\
\hline
mip-NeRF~\cite{barron2021mip} & \SetCell[r=7]{c} scene-specific & 14.62 & 0.351 & 0.495 \\
DietNeRF~\cite{jain2021putting} &  & 14.94 & 0.370 & 0.496 \\
RegNeRF~\cite{niemeyer2022regnerf} &  & 19.08 & 0.587 & 0.336 \\
MixNeRF~\cite{seo2023mixnerf} &  & 19.27 & 0.629 & \SetCell{gray!10} \\ 
FlipNeRF~\cite{seo2023flipnerf} &  & 19.34 & \SetCell{yellow!25} 0.631 & \SetCell{gray!10}\\ 
DSNeRF~\cite{deng2022depth} & \SetCell[r=2]{c} Depth guided & 18.94 & 0.582 & \SetCell{gray!10}\\ 
SparseNeRF~\cite{wang2023sparsenerf} &  & \SetCell{yellow!25} 19.86 & 0.624 & 0.328  \\
\bottomrule[1.5pt]
\end{tblr}
\end{adjustbox}
\caption{\textbf{Sparse RFF.} Scenario with 3 input views. We reproduce the values reported by~\cite{niemeyer2022regnerf} for~\cite{chibane2021stereo,yu2021pixelnerf,chen2021mvsnerf,barron2021mip,jain2021putting,niemeyer2022regnerf} and the values reported by each for~\cite{seo2023mixnerf,seo2023flipnerf,deng2022depth,wang2023sparsenerf}. We do not reproduce the LPIPS values of~\cite{seo2023mixnerf,seo2023flipnerf,deng2022depth} as they were computed using the AlexNet variant of LPIPS.}
\label{table:sparse_RFF}
\end{table}

\paragraph{\Cref{table:generalizable_DTU}: Generalizable DTU}
We reproduce the setup introduced in MVSNeRF~\cite{chen2021mvsnerf} and used in GPNR~\cite{suhail2022generalizable} on the DTU dataset~\cite{jensen2014large}.
In this setup, the images are downsampled 2$\times$ and cropped to a resolution of 640$\times$512 (images pre-processed by MVSNet~\cite{yao2018mvsnet}).
The dataset is split into 88 scenes for training with 7 lighting conditions and 16 scenes for validation\footnote{Scans [1, 8, 21, 30, 31, 34, 38, 40, 41, 45, 55, 63, 82, 103, 110, 114].}.
The images with incorrect exposure are not excluded during training. 
One scenario is considered with 10 input views, using the input/target split from~\cite{chen2021mvsnerf}.
Validation is performed on 4 views per scene\footnote{images [23, 24, 32, 44].} with lighting condition nb.~3.
We report PSNR, SSIM and LPIPS (VGG) metrics computed on masked images (foreground pixels, whose ground truth depths stand inside the scene bound). 
Our model significantly outperforms previous methods.

\begin{table}[ht]
\centering
\begin{adjustbox}{width=0.7\columnwidth}
\begin{tblr}{@{}l|c|ccc@{}}
\toprule[1.5pt]
Method & Setting & PSNR$\uparrow$ & SSIM$\uparrow$ & LPIPS$\downarrow$  \\
\hline\hline
PixelNeRF~\cite{yu2021pixelnerf} & \SetCell[r=5]{c} {generalizable\\ unknown scene} & 19.31 & 0.789 & 0.671 \\
IBRNet~\cite{wang2021ibrnet} &  & 26.04 & 0.917 & 0.190 \\
MVSNeRF~\cite{chen2021mvsnerf} &  & \SetCell{yellow!25} 26.63 & \SetCell{yellow!25} 0.931 & \SetCell{yellow!25} 0.168 \\
GPNR~\cite{suhail2022generalizable} &  & \SetCell{orange!25} 28.50 & \SetCell{orange!25}  0.932 & \SetCell{orange!25} 0.167 \\
\textbf{ConvGLR (Ours)} &  & \SetCell{red!25} 31.65 & \SetCell{red!25} 0.952 & \SetCell{red!25} 0.080 \\
\bottomrule[1.5pt]
\end{tblr}
\end{adjustbox}
\caption{\textbf{Generalizable DTU.} We reproduce the values reported by~\cite{chen2021mvsnerf} for~\cite{yu2021pixelnerf,wang2021ibrnet,chen2021mvsnerf} and the value reported by~\cite{suhail2022generalizable}. }
\label{table:generalizable_DTU}
\end{table}

\paragraph{\Cref{table:generalizable_RFF}: Generalizable RFF} 
We reproduce the setup introduced in NeRF~\cite{mildenhall2020nerf} and used in~\cite{trevithick2021grf,wang2021ibrnet,johari2022geonerf,suhail2022generalizable} on the Real Forward-Facing (RFF) dataset~\cite{mildenhall2020nerf}.
In this setup, the images are downsampled 4$\times$ to a resolution of 1008$\times$756. 
Every 8th image is used for validation, and 10 nearby input views are selected from the remaining images.
We report PSNR, SSIM and LPIPS (VGG) computed on full images.
We finetune our DTU-trained model for 50k steps on the IBRNet dataset (ConvGLR). 
We then finetune the model for another 4k steps on the 8 validation scenes (ConvGLR ft).

\begin{table}[ht]
\centering
\begin{adjustbox}{width=0.7\columnwidth}
\begin{tblr}{@{}l|c|ccc@{}}
\toprule[1.5pt]
Method & Setting & PSNR$\uparrow$ & SSIM$\uparrow$ & LPIPS$\downarrow$  \\
\hline\hline
LLFF~\cite{mildenhall2019local} & \SetCell[r=5]{c} {generalizable\\ unknown scene} & 24.13 & 0.798 & 0.212 \\
IBRNet~\cite{wang2021ibrnet} & & 25.13 & 0.817 & 0.205 \\
GeoNeRF~\cite{johari2022geonerf} & & 25.44 & 0.839 & 0.180 \\
GPNR~\cite{suhail2022generalizable} & & 25.72 & \SetCell{orange!25} 0.880 & 0.175 \\
\textbf{ConvGLR (Ours)} & & \SetCell{orange!25}  26.94 & \SetCell{yellow!25} 0.875 & \SetCell{yellow!25} 0.164 \\ 
\hline
SRN~\cite{sitzmann2019scene} & \SetCell[r=4]{c} {generalizable\\ known scene} & 22.84 & 0.668 & 0.378 \\
IBRNet ft~\cite{wang2021ibrnet} & & \SetCell{yellow!25} 26.73 & 0.851 & 0.175 \\
GeoNeRF ft 10k~\cite{johari2022geonerf} & & 26.58 & 0.856 & \SetCell{orange!25} 0.162 \\
\textbf{ConvGLR ft (Ours)} & & \SetCell{red!25} 27.81 & \SetCell{red!25} 0.889 & \SetCell{red!25} 0.125 \\
\hline
NeRF~\cite{mildenhall2020nerf} & \SetCell[r=2]{c} scene-specific & 26.50 & 0.811 & 0.250 \\
GRF~\cite{trevithick2021grf} & & 26.64 & 0.837 & 0.178 \\
\bottomrule[1.5pt]
\end{tblr}
\end{adjustbox}
\caption{\textbf{Generalizable RFF.} We reproduce the values reported by~\cite{mildenhall2020nerf} for~\cite{sitzmann2019scene,mildenhall2019local,mildenhall2020nerf} and the values reported by each for~\cite{trevithick2021grf,wang2021ibrnet,johari2022geonerf,suhail2022generalizable}. }
\label{table:generalizable_RFF}
\end{table}

\begin{table*}[ht]
\centering
\begin{adjustbox}{width=0.7\textwidth}
\begin{tblr}{@{}l|cccc|cccc|cccc@{}}
\toprule[2pt]
\SetCell[r=3]{c}{Method} & \SetCell[c=4]{c}{PSNR$\uparrow$} &  &  &  & \SetCell[c=4]{c}{SSIM$\uparrow$} &  &  &  & \SetCell[c=4]{c}{LPIPS$\downarrow$} &  &  &  \\ 
\cmidrule[lr]{2-5} \cmidrule[lr]{6-9} \cmidrule[l]{10-13}
& 12 views & \SetCell[c=3]{c} 4 views &  &  & 12 views & \SetCell[c=3]{c} 4 views &  &  & 12 views & \SetCell[c=3]{c} 4 views &  & \\
\cmidrule[lr]{2-2} \cmidrule[lr]{3-5} \cmidrule[lr]{6-6} \cmidrule[lr]{7-9} \cmidrule[lr]{10-10} \cmidrule[l]{11-13}
& dense & small & medium & large & dense & small & medium & large & dense & small & medium & large \\
\hline\hline
Soft3D~\cite{penner2017soft} & 31.93 & 30.29 & 30.84 & 30.57 & 0.940 & 0.925 & 0.930 & 0.931 & 0.052 & 0.064 & 0.060 & 0.054 \\ 
DeepView~\cite{flynn2019deepview} & \SetCell{yellow!25} 34.23 & \SetCell{yellow!25} 31.42 & \SetCell{yellow!25} 32.38 & \SetCell{yellow!25} 31.00 & \SetCell{yellow!25} 0.965 & \SetCell{yellow!25} 0.954 & \SetCell{yellow!25} 0.957 & \SetCell{yellow!25} 0.952 & \SetCell{yellow!25} 0.015 & \SetCell{yellow!25} 0.026 & \SetCell{yellow!25} 0.021 & \SetCell{yellow!25} 0.024 \\ 
MPFER~\cite{tanay2023efficient} & \SetCell{orange!25} 35.73 & \SetCell{orange!25} 33.20 & \SetCell{orange!25} 33.47 & \SetCell{orange!25} 32.38 & \SetCell{orange!25} 0.972 & \SetCell{orange!25} 0.959 & \SetCell{orange!25} 0.959 & \SetCell{orange!25} 0.953 & \SetCell{red!25} 0.012 & \SetCell{red!25} 0.018 & \SetCell{red!25} 0.018 & \SetCell{orange!25} 0.021 \\ 
\textbf{ConvGLR (Ours)} & \SetCell{red!25} 36.05  & \SetCell{red!25} 34.07 & \SetCell{red!25} 34.33 & \SetCell{red!25} 33.34 &  \SetCell{red!25} 0.977  & \SetCell{red!25} 0.968  & \SetCell{red!25} 0.968  & \SetCell{red!25} 0.964  & \SetCell{orange!25} 0.013 & \SetCell{red!25} 0.018 & \SetCell{red!25} 0.018 & \SetCell{red!25} 0.020 \\ 
\bottomrule[2pt]
\end{tblr}
\end{adjustbox}
\caption{\textbf{Spaces.} We reproduce the values reported by~\cite{tanay2023efficient} for~\cite{penner2017soft,flynn2019deepview,tanay2023efficient} (computed on images provided by the authors for~\cite{penner2017soft,flynn2019deepview}). LPIPS values were computed with the AlexNet backbone following~\cite{tanay2023efficient}.}
\label{table:Spaces}
\end{table*}

\paragraph{\Cref{table:Spaces}: Spaces.} We reproduce the setup from DeepView~\cite{flynn2019deepview} and used in MPFER~\cite{tanay2023efficient} on the Spaces dataset~\cite{flynn2019deepview}.
This dataset consists of 100 indoor and outdoor scenes, captured 5 to 10 times each using a 16-camera rig translated by small amounts.
The dataset is split into 90 scenes for training and 10 scenes for validation.
The resolution of the images is 480$\times$800.
Four scenarios are considered: one with 12 input views and three with 4 input views. Following MPFER~\cite{tanay2023efficient}, we train one model for the scenario with 12 input views and one model for the 3 scenarios with 4 input views.
Validation is performed on the first rig position for the 10 validation scenes, on the target images specified in~\cite{flynn2019deepview} for each scenario.
We report PSNR, SSIM and LPIPS (AlexNet variant) metrics computed on images after cropping an outer boundary of 16 pixels as done in~\cite{flynn2019deepview,tanay2023efficient}.
Our Convolutional Global Latent Renderer (ConvGLR) outperforms Soft3D~\cite{penner2017soft}, DeepView~\cite{flynn2019deepview} and MPFER~\cite{tanay2023efficient} by significant margins in all scenarios.

\paragraph{\Cref{table:ILSH}: ILSH} 
The Imperial Light-Stage Head dataset (ILSH)~\cite{zheng2023ilsh} was introduced as a benchmark for a recent ICCV 2023 view synthesis challenge~\cite{jang2023vschh}.
The dataset consists in 52 scenes (one individual per scene) with 24 views each at a resolution of 3000$\times$4096, with 50 views from 38 scenes held out for testing.
The dataset is publicly available upon request and blind evaluation on the test set can be performed on the Codalab platform~\cite{tonerfornottonerf}.
Evaluation is performed using PSNR and SSIM metrics, on full and masked images. 
Following the challenge organising team C1:MPFER-H~\cite{jang2023vschh}, we downsample the images 8$\times$ and train our model on the 52 scenes using 16 input views. 
Our method outperforms the challenge winner T1:OpenSpaceAI and the challenge organizing team C1:MPFER-H by more than 3dB and 1.2dB in masked PSNR respectively (metric used during the challenge).

\begin{table}[ht]
\centering
\begin{adjustbox}{width=0.7\columnwidth}
\begin{tblr}{@{}l|cc|cc|c@{}}
\toprule[1.5pt]
\SetCell[r=2]{c} Method & \SetCell[c=2]{c} PSNR$\uparrow$ & & \SetCell[c=2]{c} SSIM$\uparrow$ & & \SetCell[r=2]{c} {Time$\downarrow$ \\ (s)} \\
\cmidrule[lr]{2-3} \cmidrule[lr]{4-5}
 & full & masked & full & masked & \\
\hline\hline
C0:TensoRF & 20.54 & 26.17 & 0.71 & 0.82 & 94.02 \\
T3:CogCoVi & 21.49 & 26.33 & 0.70 & 0.82 & 806.00 \\
T2:NoNeRF & 20.37 & 26.43 & 0.69 & 0.82 & 175.58 \\
T1:OpenSpaceAI & 21.66 & 27.02 & 0.68 & \SetCell{orange!25} 0.83 & \SetCell{yellow!25} 76.88 \\
C2:DINER-SR & \SetCell{yellow!25} 22.37 & \SetCell{yellow!25} 28.50 & \SetCell{yellow!25} 0.72 & \SetCell{orange!25} 0.83 & 87.25 \\
C1:MPFER-H & \SetCell{orange!25} 28.05 & \SetCell{orange!25} 28.90 & \SetCell{orange!25} 0.84 & \SetCell{orange!25} 0.83 & \SetCell{orange!25} 1.50 \\
\textbf{ConvGLR (Ours)} & \SetCell{red!25} 28.39 & \SetCell{red!25} 30.17 & \SetCell{red!25} 0.85 & \SetCell{red!25} 0.84 & \SetCell{red!25} 0.71 \\
\bottomrule[1.5pt]
\end{tblr}
\end{adjustbox}
\caption{\textbf{ILSH dataset.} We reproduce the values from the ICCV 2023 view synthesis challenge: \emph{To NeRF or not to NeRF}~\cite{jang2023vschh}.}
\label{table:ILSH}
\end{table}

\paragraph{\Cref{table:ablation}: Ablations}
We perform ablations on the Sparse DTU setup with 9 input views, and train each model for 50k steps on patches of 256$\times$256 pixels.
We start by training our full model (line~10).
We then consider 3 variants of our backbone architecture. 
\emph{No PSV}: the input images are concatenated and processed as a group, but no PSV is constructed (line~1). 
\emph{MVS-based}: deep features are extracted from individual input images and a cost volume is constructed by computing the variance over the views (line~2).
\emph{MPI-based}: the model outputs $D$ RGB$\alpha$ images that are then alpha-blended (line~3).
We see that our ConvGLR backbone produces the best results by big margins, validating our choice of a PSV based architecture rendering novel views globally in a low-dimensional latent space.
We then train the same model 4 times, on image patches ranging from 16$\times$16 to 128$\times$128 (lines~4-7). 
In order to keep the effective batch size constant, we train on $256\times256$ patches that we slice into $16^2$, $8^2$, $4^2$ and $2^2$ pieces respectively.
We see that the performance degrades sharply with smaller patch sizes, confirming that the global rendering contributes significantly to the performance of our approach.
Finally we turn off the positional and angular encodings together and separately (lines~8-9).
We see that both contribute to the final performance of the model.

\begin{table}[ht]
\centering
\begin{adjustbox}{width=\columnwidth}
\begin{tblr}{@{}c|c|c|c|c|c|c|c@{}}
\toprule[1.5pt]
line & pos. enc. & ang. enc. & backbone & patch size & params & FLOPS & PSNR$\uparrow$ \\ 
\hline\hline
1 & \textbf{Yes} & \textbf{Yes} & No PSV & \textbf{256$\times$256} & 29.6M & 0.3T & 17.30 \\ 
2 & \textbf{Yes} & \textbf{Yes} & MVS-based & \textbf{256$\times$256} & 40.1M  & 6.1T & 23.39 \\ 
3 & \textbf{Yes} & \textbf{Yes} & MPI-based & \textbf{256$\times$256} & 28.2M & 7.8T & 24.62 \\ 
4 & \textbf{Yes} & \textbf{Yes} & \textbf{ConvGLR} & 16$\times$16 & 40.3M & 6.6T & 24.03 \\ 
5 & \textbf{Yes} & \textbf{Yes} & \textbf{ConvGLR} & 32$\times$32 & 40.3M & 6.6T & 25.79 \\ 
6 & \textbf{Yes} & \textbf{Yes} & \textbf{ConvGLR} & 64$\times$64 & 40.3M & 6.6T & \SetCell{orange!25} 26.22 \\ 
7 & \textbf{Yes} & \textbf{Yes} & \textbf{ConvGLR} & 128$\times$128 & 40.3M & 6.6T & \SetCell{yellow!25} 26.20 \\ 
8 & No & No & \textbf{ConvGLR} & \textbf{256$\times$256} & 40.2M & 6.5T & 25.66 \\
9 & \textbf{Yes} & No & \textbf{ConvGLR} & \textbf{256$\times$256} & 40.3M & 6.5T & 25.91\\ 
10 & \textbf{Yes} & \textbf{Yes} & \textbf{ConvGLR} & \textbf{256$\times$256} & 40.3M & 6.6T & \SetCell{red!25} 26.33 \\ 
\bottomrule[1.5pt]
\end{tblr}
\end{adjustbox}
\caption{\textbf{Ablations.} All the models were trained on the Sparse DTU setup with 9 input views for 50k steps.}
\label{table:ablation}
\end{table}

\section{Conclusion}

We introduced global latent neural rendering, a novel view synthesis approach that consists in learning a generalizable light field model from plane sweep volumes, 
and ConvGLR, a convolutional architecture that implements this idea efficiently.
While ConvGLR performs remarkably well, we believe that there is still room for improvement by optimizing the architecture, scaling up the training, and sampling the depth planes in a scene-adaptive manner.

{
    \small
    \bibliographystyle{ieeenat_fullname}
    \bibliography{main}
}

\clearpage
\maketitlesupplementary

\section{Plane sweep volumes}

\begin{figure*}[p]
  \centering
  \includegraphics[width=\linewidth]{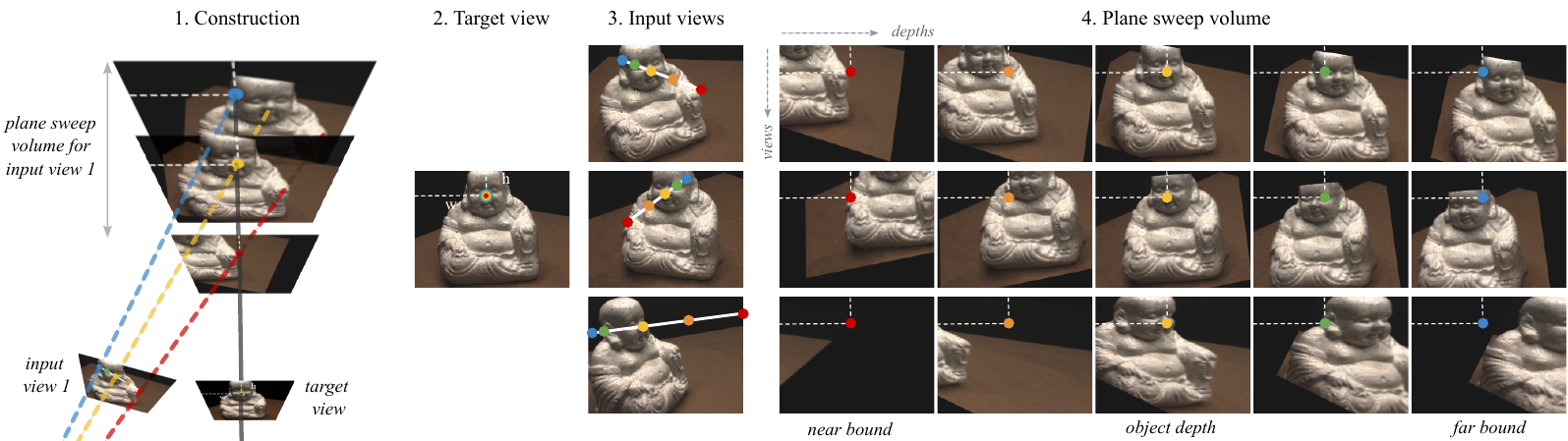}

   \caption{\textbf{The epipolar geometry of the plane sweep volume.} 1. The PSV is constructed by projecting each input view on a set of planes distributed parallel to the target image plane. 2. The camera ray passing through the pixel location (h, w) in the target image plane (gray line in 1.) projects as a set of epipolar lines in the input views (white lines in 3.). 4. Moving along the depth dimension of the PSV at pixel location (h, w) is equivalent to moving along the corresponding epipolar lines for each input view. The actual depth of the object at pixel location (h, w) is found when the local image features match across views (yellow dot).}
   \label{fig:epipolar_geometry}
\end{figure*}

\begin{figure*}[p]
  \centering
  \includegraphics[width=0.8\linewidth]{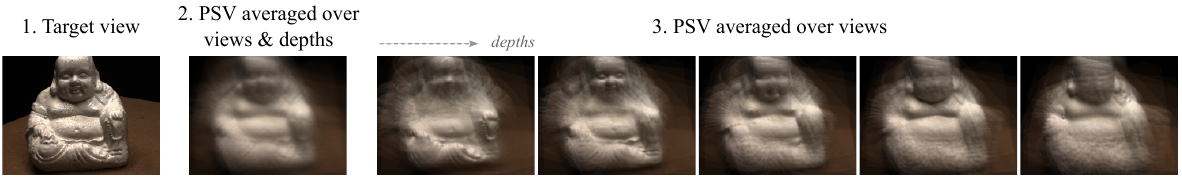}
   \caption{\textbf{Averaging the plane sweep volume.} 1. The target view for which a plane sweep volume is constructed, using 9 input views (not including the target view) and \emph{near} and \emph{far} bounds that are close to the object depth.
   2. Averaging the PSV over views and depths provides a blurry estimate of the target views. 3. Averaging the PSV over the views brings successive depths of the object into focus.}
   \label{fig:PSV_average}
\end{figure*}

As discussed in \cref{sec:method}, the plane sweep volume is a highly structured tensor encoding the epipolar geometry between the input views and the target view. We describe this epipolar geometry in more details in \cref{fig:epipolar_geometry}.
One of the interesting properties of the PSV is that local image features match across the input views when a depth plane is precisely located on an object in the scene. 
A simple way to highlight this property is to average the PSV over the input views, as done in \cref{fig:PSV_average}.
There, each depth plane slices the 3D object at a specific depth. When a part of the object is located on the depth plane, this part appears ``in focus'' in the mean PSV. 
On the contrary, the parts that are located at other depths appear blurry and out of focus.
Such averaging of the PSV is closely related to the original plane sweep algorithm of Collins~\cite{collins1996space} for depth estimation, and further motivates the use of plane sweep volumes for novel view synthesis.

\section{Implementation details}

We presented an overview of our Convolutional Global Latent Renderer (ConvGLR) in \cref{sec:method} and \cref{fig:ConvGLR_overview} of the main paper.  
ConvGLR transforms 5D input PSVs into 3D rendered images in 4 steps: (1) Grouped PSV, (2) Multi-view matching, (3) global latent rendering and (4) upsampling.
We provide more details in \cref{table:convGLR} where all the operations are listed with their effect on the dimension of the input tensor.
Particular emphasis has been put on memory efficiency and in-place viewing operations are used extensively while expensive reshape or transpose operations are avoided.

We propose two possible implementations of the global latent rendering step: 
one where the resblocks are applied over the depths with shared weights by using the batch dimension for parallel processing,
and one where the resblocks are applied over the depths with specialized weights by moving the depths into the channel dimension and applying resblocks implemented with grouped convolutions.
In practice, we did not observe any significant difference of performance between the two implementations.

\section{Qualitative results}

\begin{figure*}[ht]
  \centering
    \makebox[0.32\textwidth]{RegNeRF~\cite{niemeyer2022regnerf}}\hfill
    \makebox[0.32\textwidth]{ConvGLR (Ours)}\hfill
    \makebox[0.32\textwidth]{Ground Truth}

    \includegraphics[width=0.32\textwidth]{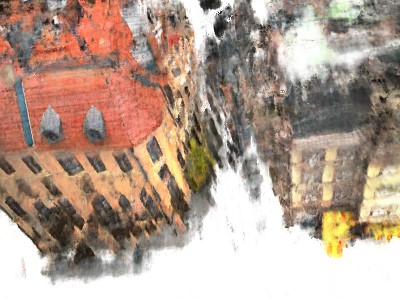}\hfill
    \includegraphics[width=0.32\textwidth]{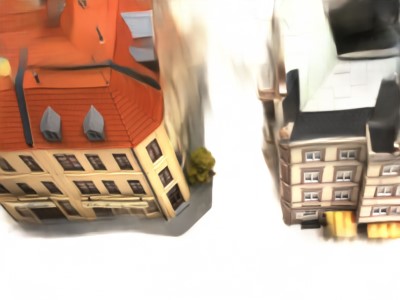}\hfill
    \includegraphics[width=0.32\textwidth]{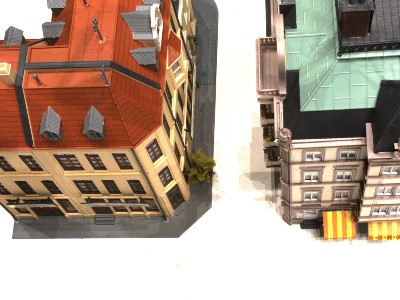}

    \includegraphics[width=0.32\textwidth]{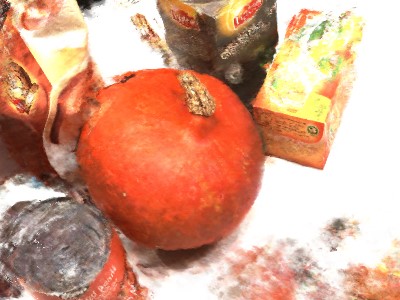}\hfill
    \includegraphics[width=0.32\textwidth]{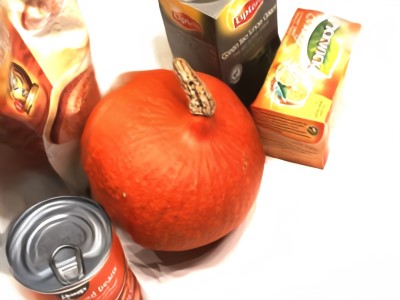}\hfill
    \includegraphics[width=0.32\textwidth]{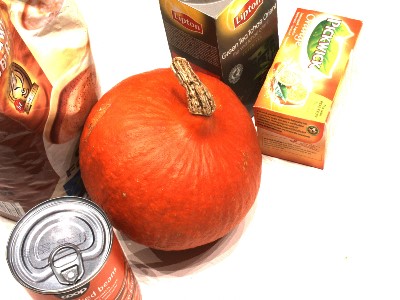}

   \caption{\textbf{Qualitative results.} Sparse DTU.}
   \label{fig:qualitative_results1}
\end{figure*}

\begin{figure*}[ht]
  \centering
    \makebox[0.32\textwidth]{RegNeRF~\cite{niemeyer2022regnerf}}\hfill
    \makebox[0.32\textwidth]{ConvGLR (Ours)}\hfill
    \makebox[0.32\textwidth]{Ground Truth}

    \includegraphics[width=0.32\textwidth]{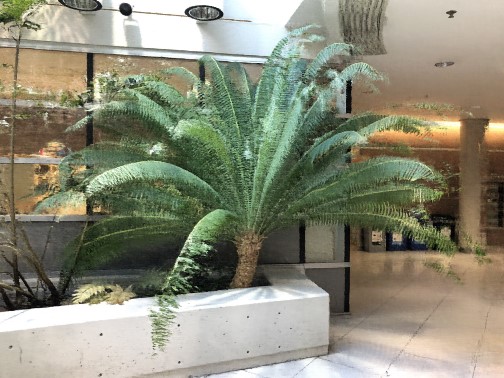}\hfill
    \includegraphics[width=0.32\textwidth]{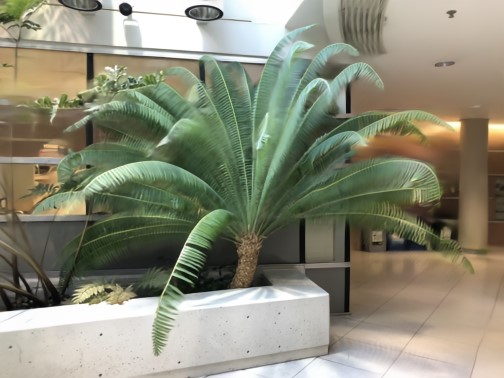}\hfill
    \includegraphics[width=0.32\textwidth]{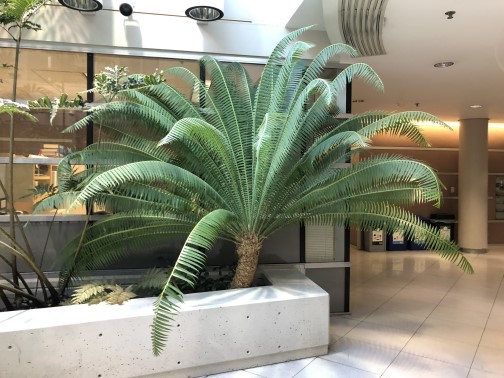}

    \includegraphics[width=0.32\textwidth]{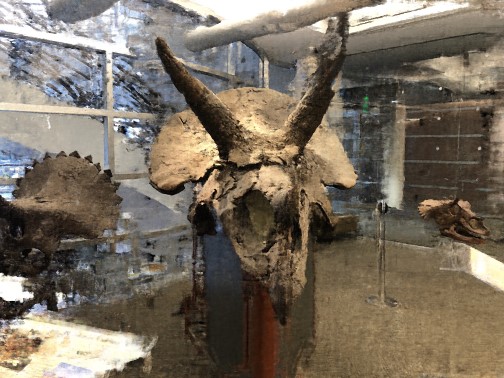}\hfill
    \includegraphics[width=0.32\textwidth]{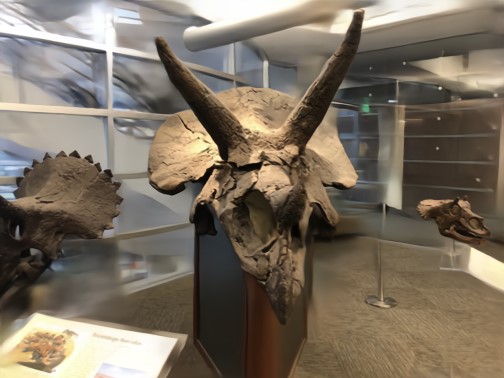}\hfill
    \includegraphics[width=0.32\textwidth]{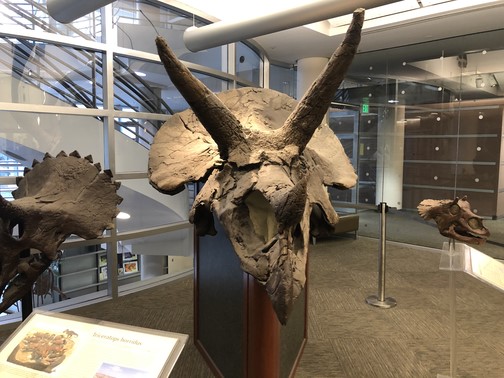}

   \caption{\textbf{Qualitative results.} Sparse RFF.}
   \label{fig:qualitative_results2}
\end{figure*}

\begin{figure*}[ht]
  \centering
    \makebox[0.32\textwidth]{GPNR~\cite{suhail2022generalizable}}\hfill
    \makebox[0.32\textwidth]{ConvGLR (Ours)}\hfill
    \makebox[0.32\textwidth]{Ground Truth}

    \includegraphics[width=0.32\textwidth]{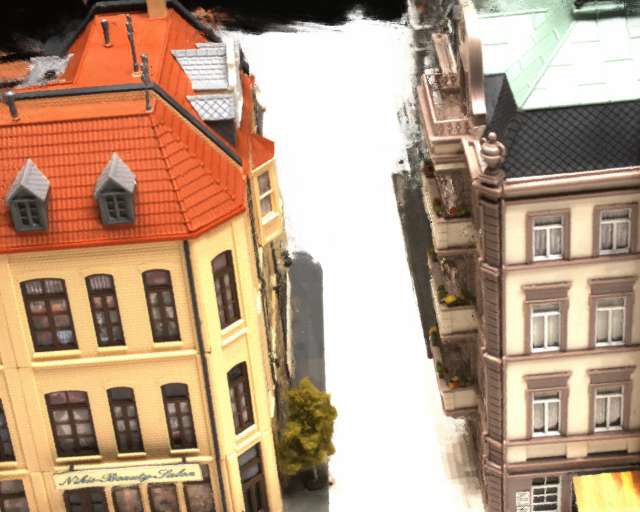}\hfill
    \includegraphics[width=0.32\textwidth]{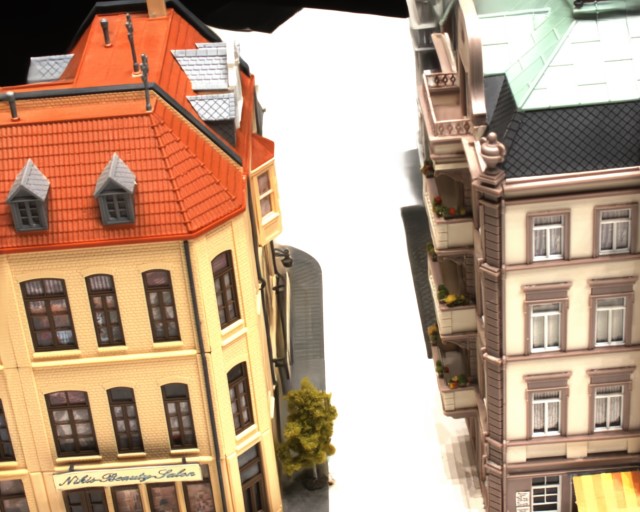}\hfill
    \includegraphics[width=0.32\textwidth]{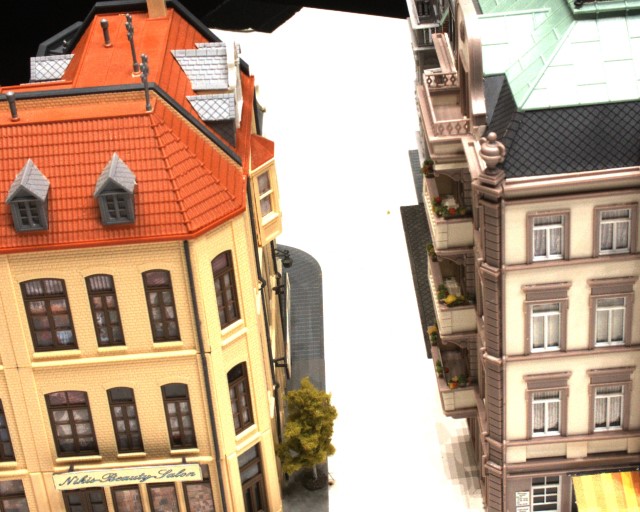}

    \includegraphics[width=0.32\textwidth]{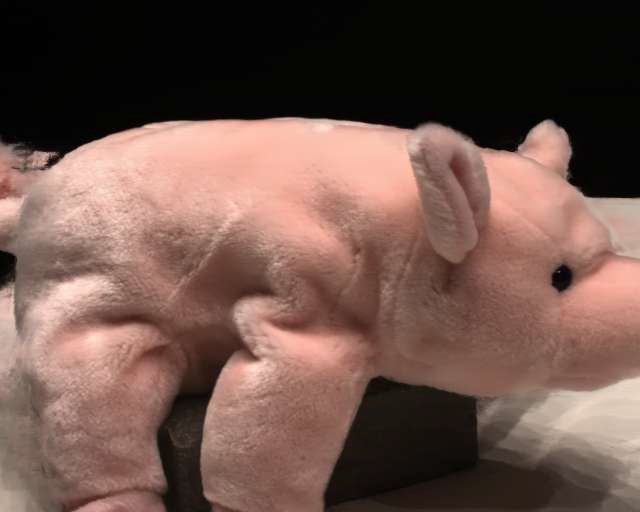}\hfill
    \includegraphics[width=0.32\textwidth]{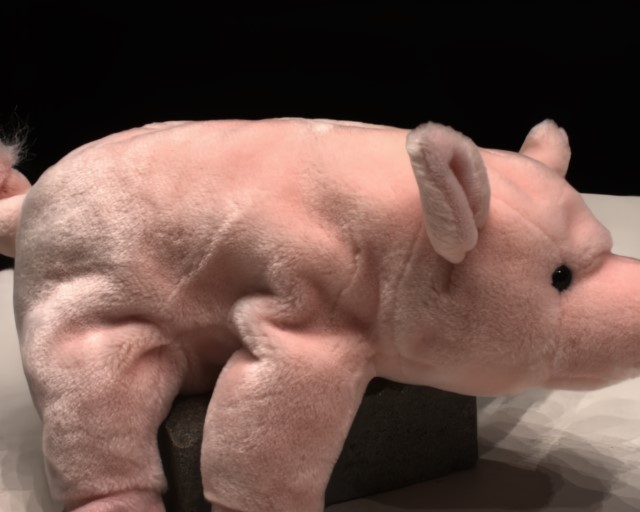}\hfill
    \includegraphics[width=0.32\textwidth]{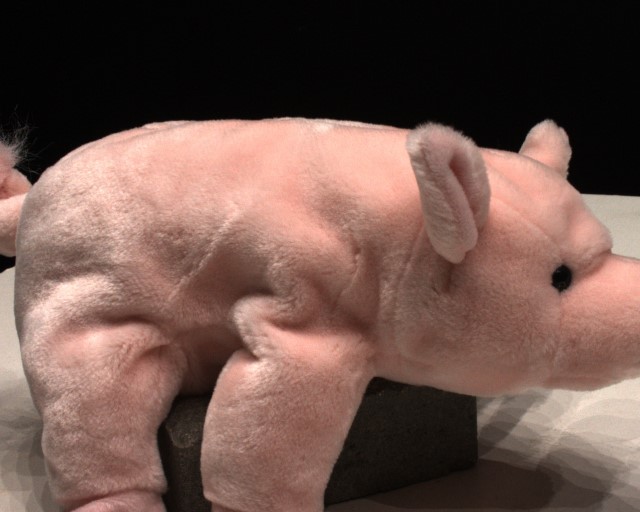}

   \caption{\textbf{Qualitative results.} Generalizable DTU (unknown scenes).}
   \label{fig:qualitative_results3}
\end{figure*}

\begin{figure*}[ht]
  \centering
    \makebox[0.32\textwidth]{GeoNeRF~\cite{johari2022geonerf}}\hfill
    \makebox[0.32\textwidth]{ConvGLR (Ours)}\hfill
    \makebox[0.32\textwidth]{Ground Truth}

    \includegraphics[width=0.32\textwidth]{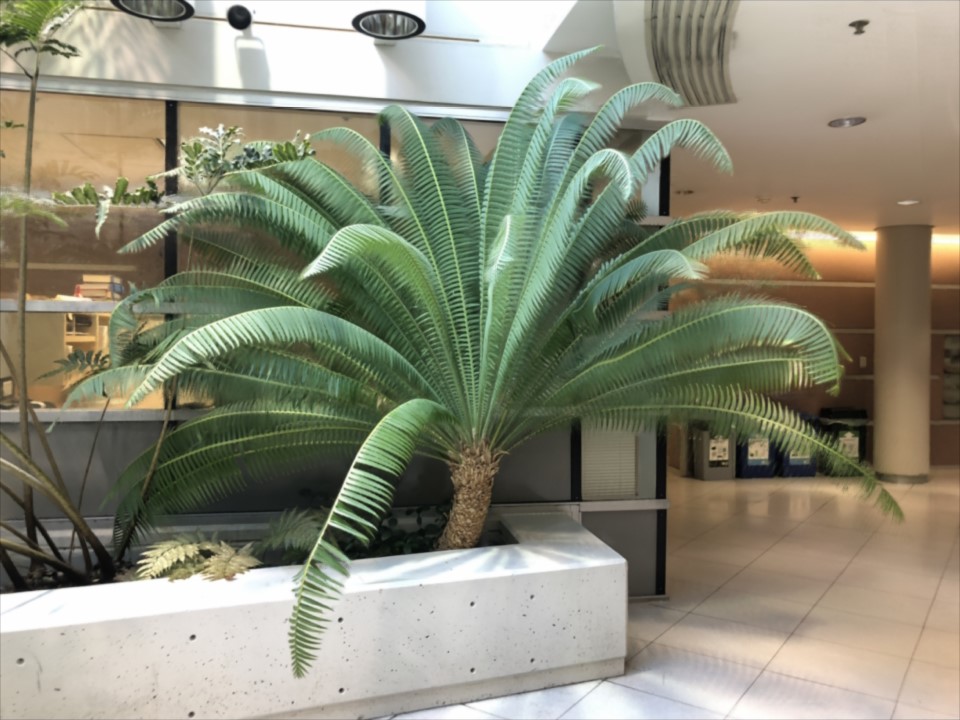}\hfill
    \includegraphics[width=0.32\textwidth]{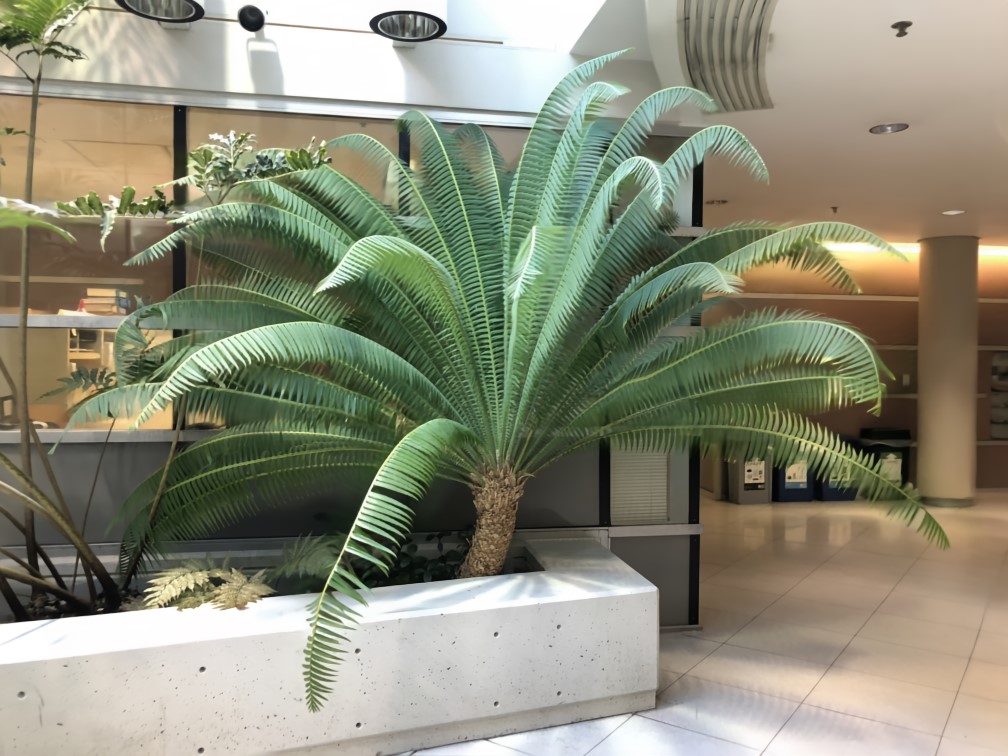}\hfill
    \includegraphics[width=0.32\textwidth]{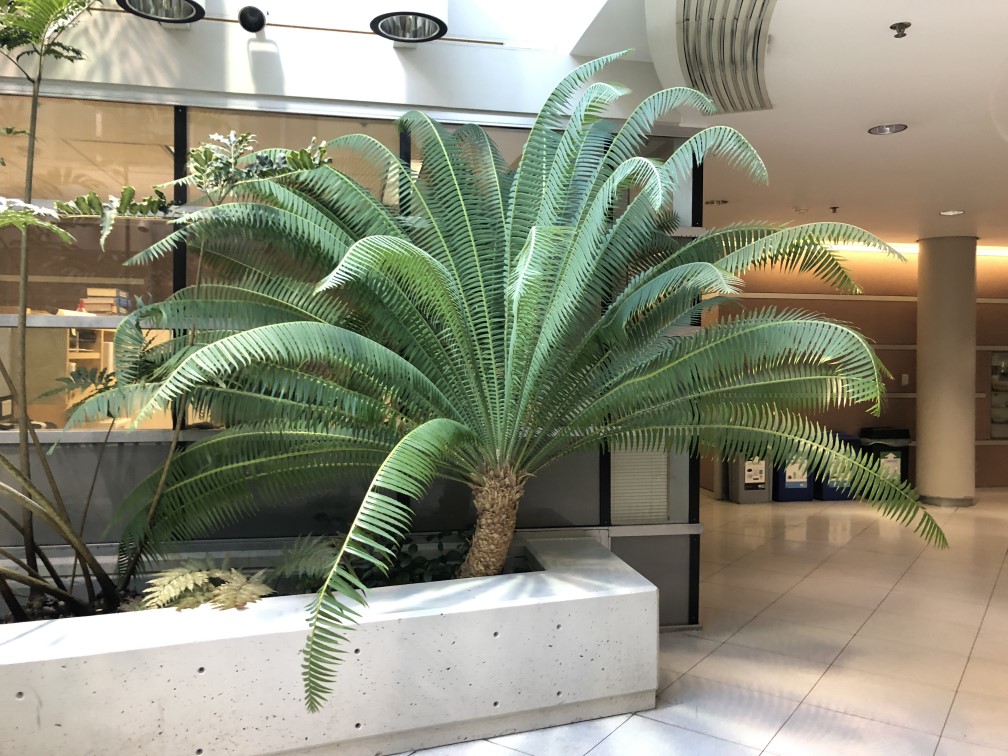}

    \includegraphics[width=0.32\textwidth]{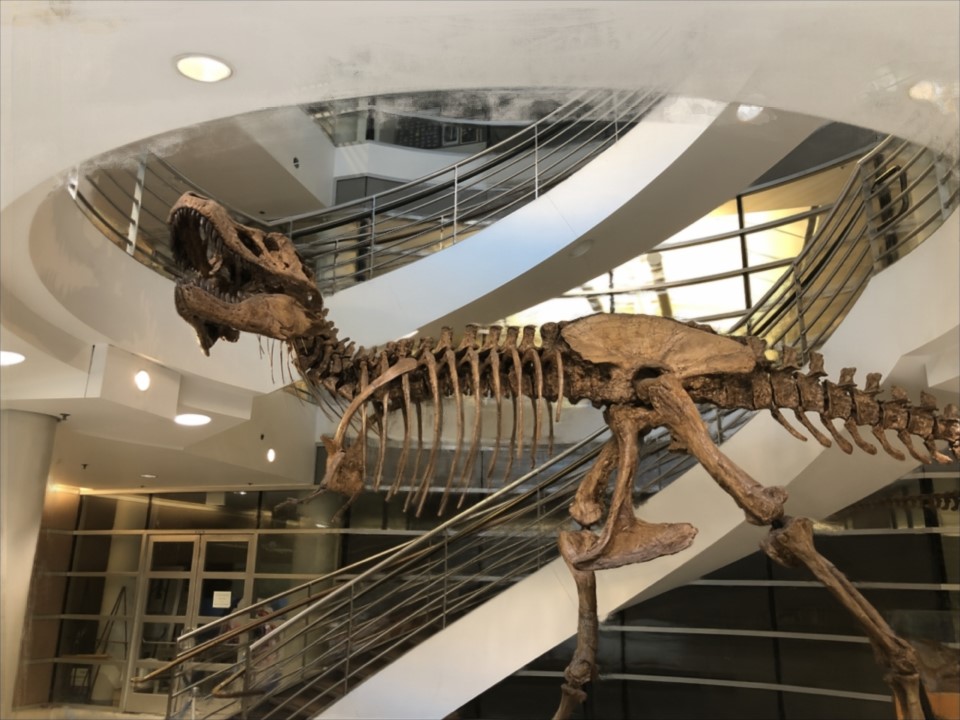}\hfill
    \includegraphics[width=0.32\textwidth]{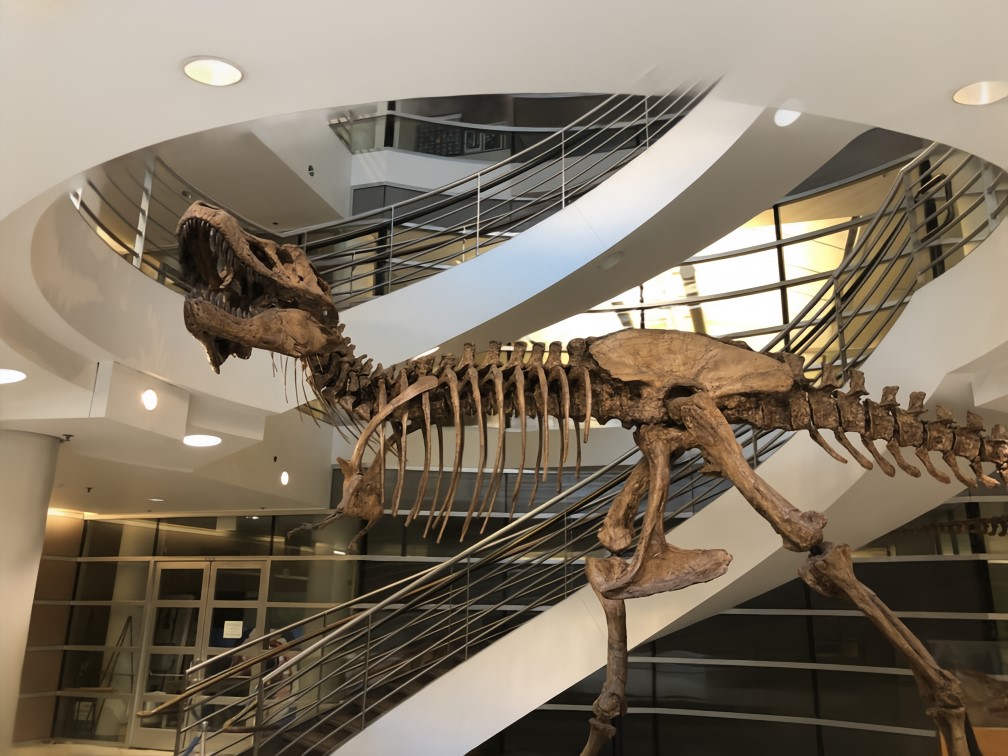}\hfill
    \includegraphics[width=0.32\textwidth]{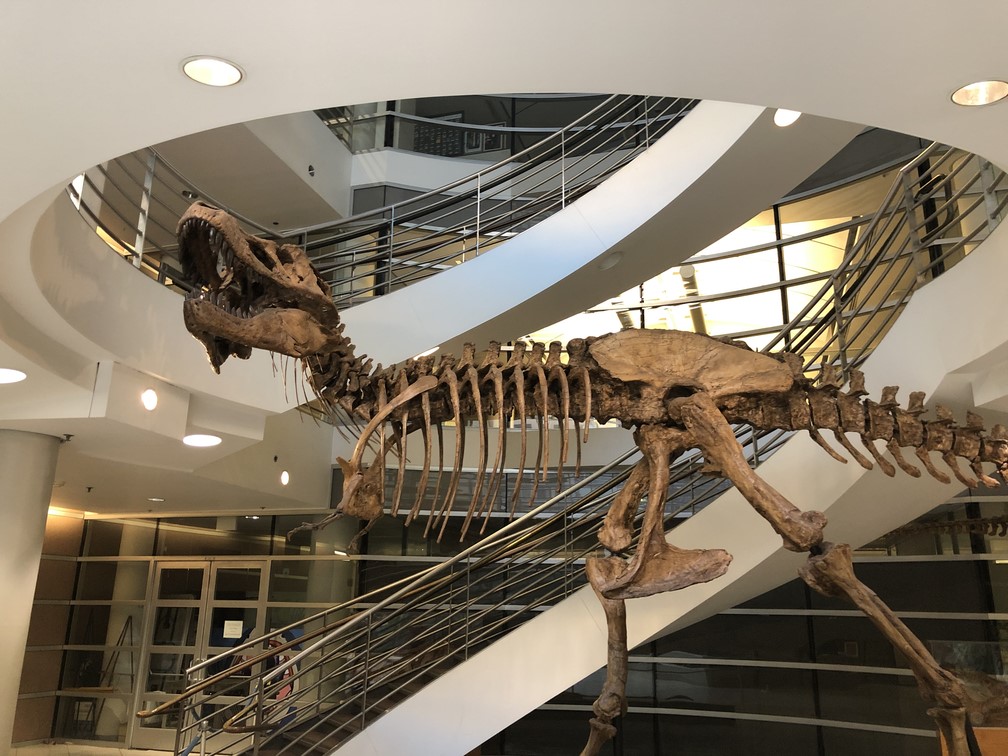}
   \caption{\textbf{Qualitative results.} Generalizable RFF (unknown scenes).}
   \label{fig:qualitative_results4}
\end{figure*}

We provide a number of qualitative comparisons to baselines in \cref{fig:qualitative_results1}, \cref{fig:qualitative_results2}, \cref{fig:qualitative_results3}, \cref{fig:qualitative_results4}.

\begin{table*}[p]
\centering
\begin{adjustbox}{width=0.8\textwidth}
\begin{tblr}{@{}l|lcccc|lcccc@{}}
\toprule[2pt]
implementation & \SetCell[c=5]{c} shared weights & & & & & \SetCell[c=5]{c} specialized weights & & & & \\
\hline
\SetCell[r=2]{c} block & \SetCell[r=2]{c} description & \SetCell[c=4]{c} output dimension &  &  &  & \SetCell[r=2]{c} description & \SetCell[c=4]{c} output dimension &  &  & \\
\cmidrule[lr]{3-6} \cmidrule[lr]{8-11}
 &  & batch & channels & height & width &  & batch & channels & height & width \\
\hline\hline
\SetCell[r=3]{c} {grouped\\ PSV} & 5D PSV & $D \; V$ & $3$ & $H$ & $W$ & \SetCell[r=3,c=5]{gray!25}\\
  & concatenate views & $D$  & $3V$ & $H$ & $W$ \\
  & view & $D_{\scriptscriptstyle{\!G}}$ & $3GV$ & $H$ & $W$ \\
\hline
\SetCell[r=6]{c} {multi-view\\ matching} & \textbf{conv.} & $D_{\scriptscriptstyle{\!G}}$ & $C$ & $H$ & $W$ & \SetCell[r=6,c=5]{gray!25}\\
  & \textbf{2 resblocks} & $D_{\scriptscriptstyle{\!G}}$ & $C$ & $H$ & $W$ \\
  & \textbf{conv.} (stride 2) & $D_{\scriptscriptstyle{\!G}}$ & $2C$ & $H/2$ & $W/2$ \\
  & \textbf{3 resblocks} & $D_{\scriptscriptstyle{\!G}}$ & $2C$ & $H/2$ & $W/2$ \\
  & \textbf{conv.} (stride 2) & $D_{\scriptscriptstyle{\!G}}$ & $4C$ & $H/4$ & $W/4$ \\
  & \textbf{4 resblocks} & $D_{\scriptscriptstyle{\!G}}$ & $4C$ & $H/4$ & $W/4$ \\
\hline
\SetCell[r=10]{c} {global latent\\ rendering} & view & $D_{\scriptscriptstyle{\!G}}/2$ & $8C$ & $H/4$ & $W/4$ & view & $1$ & $D_{\scriptscriptstyle{\!G}}\!\times\!4C$ & $H/4$ & $W/4$ \\
  & \textbf{1 resblock} & $D_{\scriptscriptstyle{\!G}}/2$ & $4C$ & $H/4$ & $W/4$ & \SetCell[r=2]{l} {\textbf{1 resblock}\\ ($D_{\scriptscriptstyle{\!G}}/2$ groups)} & \SetCell[r=2]{c} $1$ & \SetCell[r=2]{c} $D_{\scriptscriptstyle{\!G}}/2\!\times\!4C$ & \SetCell[r=2]{c} $H/4$ & \SetCell[r=2]{c} $W/4$ \\
  & view & $D_{\scriptscriptstyle{\!G}}/4$ & $8C$ & $H/4$ & $W/4$ \\
  & \textbf{1 resblock} & $D_{\scriptscriptstyle{\!G}}/4$ & $4C$ & $H/4$ & $W/4$ & \SetCell[r=2]{l} {\textbf{1 resblock}\\ ($D_{\scriptscriptstyle{\!G}}/4$ groups)} & \SetCell[r=2]{c} $1$ & \SetCell[r=2]{c} $D_{\scriptscriptstyle{\!G}}/4\!\times\!4C$ & \SetCell[r=2]{c} $H/4$ & \SetCell[r=2]{c} $W/4$ \\
  & view & $D_{\scriptscriptstyle{\!G}}/8$ & $8C$ & $H/4$ & $W/4$ \\
  & \textbf{1 resblock} & $D_{\scriptscriptstyle{\!G}}/8$ & $4C$ & $H/4$ & $W/4$ & \SetCell[r=2]{l} {\textbf{1 resblock}\\ ($D_{\scriptscriptstyle{\!G}}/8$ groups)} & \SetCell[r=2]{c} $1$ & \SetCell[r=2]{c} $D_{\scriptscriptstyle{\!G}}/8\!\times\!4C$ & \SetCell[r=2]{c} $H/4$ & \SetCell[r=2]{c} $W/4$ \\
  & view & $D_{\scriptscriptstyle{\!G}}/16$ & $8C$ & $H/4$ & $W/4$ \\
  & \textbf{1 resblock} & $D_{\scriptscriptstyle{\!G}}/16$ & $4C$ & $H/4$ & $W/4$ & \SetCell[r=2]{l} {\textbf{1 resblock}\\ ($D_{\scriptscriptstyle{\!G}}/16$ groups)} & \SetCell[r=2]{c} $1$ & \SetCell[r=2]{c} $D_{\scriptscriptstyle{\!G}}/16\!\times\!4C$ & \SetCell[r=2]{c} $H/4$ & \SetCell[r=2]{c} $W/4$ \\
  & view & 1 & $D_{\scriptscriptstyle{\!G}}/16\!\times\!4C$ & $H/4$ & $W/4$ \\
  & \textbf{1 resblock} & $1$ & $4C$ & $H/4$ & $W/4$ & \textbf{1 resblock} & $1$ & $4C$ & $H/4$ & $W/4$ \\
\hline
\SetCell[r=5]{c} {upsampling} & interpolate (nearest) & $1$ & $4C$ & $H/2$ & $W/2$ & \SetCell[r=5,c=5]{gray!25} \\
  & \textbf{3 resblocks} & $1$ & $2C$ & $H/2$ & $W/2$ \\
  & interpolate (nearest) & $1$ & $2C$ & $H$ & $W$ \\
  & \textbf{2 resblocks} & $1$ & $C$ & $H$ & $W$ \\
  & \textbf{conv.} & $1$ & $3$ & $H$ & $W$ \\
\bottomrule[2pt]
\end{tblr}
\end{adjustbox}
\caption{\textbf{ConvGLR.} The 5D plane sweep volume is progressively turned into a 3D rendered image by applying a succession of 2D convolutions and resblocks while making effective use of viewing operations and batching. Learnable blocks are emphasized in bold.}
\label{table:convGLR}
\end{table*}

\end{document}